\documentclass[10pt,twocolumn,letterpaper]{article}

\usepackage{cvpr}
\usepackage{times}
\usepackage{epsfig}
\usepackage{graphicx}
\usepackage{amsmath}
\usepackage{amssymb}
\usepackage{subfig}
\usepackage{booktabs}
\usepackage{pdfpages}
\DeclareMathOperator*{\argmin}{arg\,min}
\DeclareMathOperator*{\argmax}{arg\,max}

\usepackage[pagebackref=true,breaklinks=true,letterpaper=true,colorlinks,bookmarks=false]{hyperref}

\cvprfinalcopy 
\ifcvprfinal\fi

\begin{document}

\graphicspath{{Images/}}
\DeclareGraphicsExtensions{.png,.pdf,.jpg,.mps,.jpeg,.jbig2,.jb2,.JPG,.JPEG,.JBIG2,.JB2}
\newcommand{\comment}[2]{#2}
\title{Lifting from the Deep: Convolutional 3D Pose
Estimation from a Single Image} 

\author{Denis Tome               \\
University College London                       \\
\href{mailto:D.Tome@cs.ucl.ac.uk}{D.Tome@cs.ucl.ac.uk}
\and
Chris Russell                      \\
 The Turing Institute and\\
The University of Edinburgh\\
\href{mailto:crussell@turing.ac.uk}{crussell@turing.ac.uk}
\and
Lourdes Agapito\\
University College London                       \\
\href{mailto:l.agapito@cs.ucl.ac.uk}{l.agapito@cs.ucl.ac.uk}
\and
\url{http://visual.cs.ucl.ac.uk/pubs/liftingFromTheDeep}
}

\maketitle
\thispagestyle{empty}

\begin{abstract}
  We propose a unified formulation for the problem of 3D human pose
  estimation from a single raw RGB image that reasons jointly about 2D
  joint estimation and 3D pose reconstruction to improve both tasks.
  We take an integrated approach that fuses probabilistic knowledge of
  3D human pose with a multi-stage CNN architecture and uses the
  knowledge of plausible 3D landmark locations to refine the search
  for better 2D locations. The entire process is trained end-to-end,
  is extremely efficient and obtains state-of-the-art results on
  Human3.6M outperforming previous approaches both on 2D and 3D
  errors.
  \vspace{-5mm}
\end{abstract}

\section{Introduction}
Estimating the full 3D pose of a human from a single RGB image is one
of the most challenging problems in computer vision. It involves
tackling two inherently ambiguous tasks. First, the 2D location of the
human joints, or landmarks, must be found in the image, a problem
plagued with ambiguities due to the large variations in visual
appearance caused by different camera viewpoints, external and self
occlusions or changes in clothing, body shape or illumination. Next,
lifting the coordinates of the 2D landmarks into 3D from a single
image is still an ill-posed problem -- the space of possible 3D poses
consistent with the 2D landmark locations of a
human, 
is infinite. Finding the 
correct 3D pose that matches the image requires injecting additional
information usually in the form of 3D geometric pose priors and
temporal or structural constraints.


We propose a new joint approach to 2D landmark detection and full 3D
pose estimation from a single RGB image that takes advantage of
reasoning jointly about the estimation of 2D and 3D landmark locations
to improve both tasks.  We propose a novel CNN architecture that
learns to combine the image appearance based predictions provided by
\emph{convolutional-pose-machine} style 2D landmark
detectors~\cite{wei2016convolutional}, with the geometric 3D skeletal
information encoded in a novel pretrained model of 3D human pose. %

Information captured by the 3D human pose model is embedded in the 
CNN architecture as an additional layer that
lifts 2D landmark coordinates into 3D while imposing that they lie on the
space of physically plausible poses. The advantage of integrating the
output proposed by the 2D landmark location predictors -- based purely on
image appearance -- with the 3D pose predicted by a probabilistic model,
is that the 2D landmark location estimates are improved by guaranteeing
that they satisfy the anatomical 3D constraints encapsulated in the
human 3D pose model. In this way, both tasks clearly benefit from each
other.

\begin{figure*}\vspace{-7mm}
\begin{center}
\includegraphics[width=\linewidth]{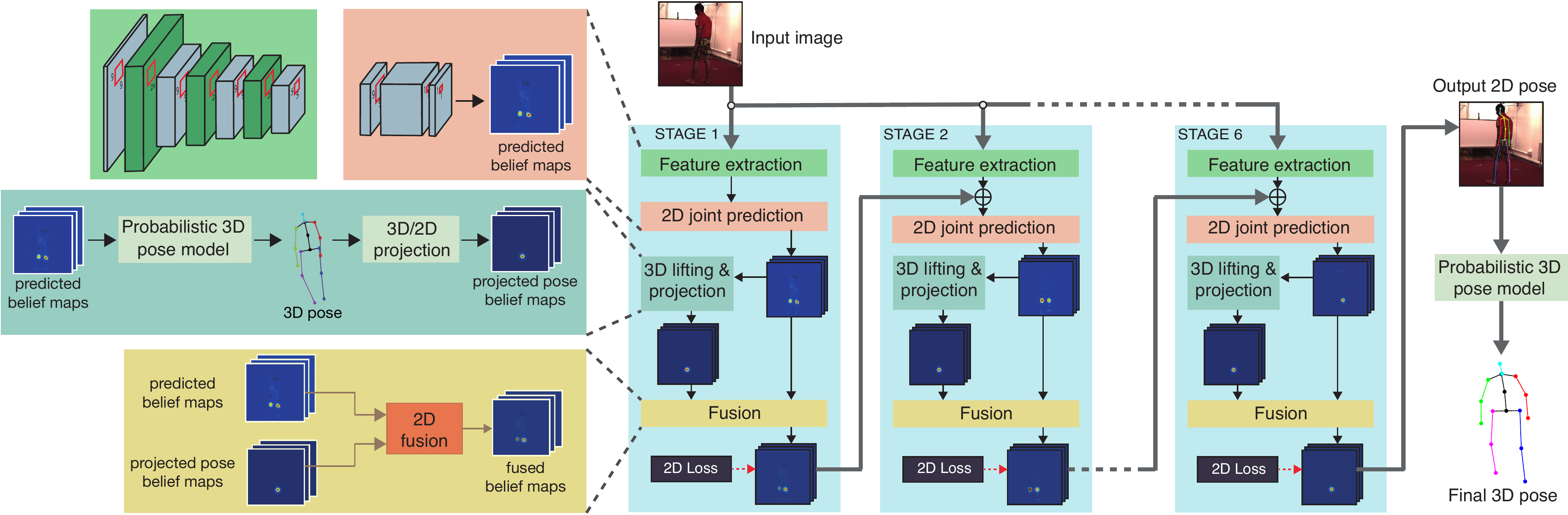}
\end{center}
\vspace{-7mm}
\caption{\small The multistage deep architecture for 2D/3D human pose
  estimation. Each stage produces as output a set of belief maps
  for the location of the 2D landmarks (one per landmark). The belief
  maps from each stage, as well as the image, are used as input to
  the next stage. Internally, each stage learns to combine: \emph{(a)}
  belief maps provided by convolutional 2D joint predictors, with
  \emph{(b)} projected pose belief maps, proposed by the probabilistic
  3D pose model. 
     The 3D pose layer is responsible for lifting 2D landmark
     coordinates into 3D and projecting them onto the space of valid
     3D poses.  These two belief maps are then fused into a single set
     of output proposals for the 2D landmark locations per stage.  The
     accuracy of the 2D and 3D landmark locations increases
     progressively through the stages. The loss used at each stage requires only 2D pose annotations, not 3D. The overall architecture is
     fully differentiable -- including the new projected-pose belief
     maps and 2D-fusion layers -- and can be
     trained end-to-end using
     back-propagation. [Best viewed in color.] \label{fig:pipeline}}\vspace{-5mm}
\end{figure*}


A further advantage of our approach is that the  2D and 3D training
data sources may be completely independent. The  deep
architecture only needs that images are annotated with 2D poses, not
3D poses. The human pose model is trained independently and
exclusively from 3D mocap data. This decoupling between 2D and 3D
training data presents a huge advantage since we can augment the 
training sets completely independently. For instance we can take
advantage of extra 2D pose annotations without the need for 3D ground
truth or extend the 3D training data to further mocap datasets without
the need for synchronized 2D images.

\noindent{\bf Our contribution:} In this work, we show how to integrate a
prelearned 3D human pose model directly within a novel CNN architecture
 (illustrated in figure~\ref{fig:pipeline}) for joint 2D landmark
and 3D human pose estimation. In contrast to preexisting methods, we
do not take a pipeline approach that takes 2D landmarks as
given. Instead, we show how such a model can be used as part of the
CNN architecture itself, and how the architecture can learn to use
physically plausible 3D reconstructions in its search for better 2D
landmark locations. Our method achieves state-of-the-art results on
the Human3.6M dataset both in terms of 2D and 3D errors.


\section{Related Work}
%
We first describe methods that assume that 2D joint
locations are provided as input and focus on solving the 3D lifting
problem and follow with methods that learn to estimate the 3D pose
directly from images.

\noindent{\bf 3D pose from known 2D joint positions:} A large body of work
has focused on recovering the 3D pose of people given perfect 2D joint
positions as input. Early approaches \cite{lee1985determination,
  taylor2000reconstruction, parameswaran2004view,
  barron2001estimating} took advantage of anatomical knowledge of the
human skeleton or joint angle limits to recover pose from a single
image. More recent methods \cite{fan2014pose,
  ramakrishna2012reconstructing,akhter2015pose} have focused on learning a prior
statistical model of the human body directly from 3D mocap
data.

Non-rigid structure from motion approaches (NRSfM) also recover 3D
articulated
motion~\cite{bregler2000recovering,Akhter:etal:PAMI:2011,Gotardo:Martinez:PAMI:2012,Lee:etal:PAMI}
given known 2D correspondences for the joints in every frame of a
monocular video. Their huge advantage, as unsupervised methods, is
they do not need 3D training data, instead they can learn a linear
basis for the 3D poses purely from 2D data.
Their main drawback is their need for significant camera movement
throughout the sequence to guarantee accurate 3D
reconstruction.  Recent work on NRSfM applied to human pose estimation has
focused on escaping these limitations by the use of a linear
model to represent shape variations of the human body. For
instance,~\cite{cho2016complex} defined a generative
model based on the assumption that complex shape variations can be
decomposed into a mixture of primitive shape variations and achieve
competitive results.

Representing human 3D pose as a linear combination of a sparse set of
3D bases, pretrained using 3D mocap data, has also proved a popular
approach for articulated human
motion~\cite{ramakrishna2012reconstructing,
  Wang:etal:CVPR:2014,zhou2015sparse}, while~\cite{zhou2015sparse}
propose a convex relaxation to jointly estimate the coefficients of
the sparse representation and the camera
viewpoint~\cite{ramakrishna2012reconstructing}
and~\cite{Wang:etal:CVPR:2014} enforce limb length
constraints. Although these approaches can reconstruct 3D pose from a
single image, their best results come from imposing temporal
smoothness on the reconstructions of a video sequence.

Recently, Zhao \etal~\cite{zhao2016simple} achieved state-of-the-art
results by training a simple neural network to recover 3D pose from
known 2D joint positions. Although the results on perfect 2D input data
are impressive, the inaccuracies in 2D joint estimation are not
modeled and the performance of this approach combined with joint
detectors is unknown.

\noindent{\bf 3D pose from images:} Most approaches to 3D pose
inference directly from images fall into one of two categories:
\emph{(i)} models that learn to regress the 3D pose directly from
image features and \emph{(ii)} pipeline approaches where the 2D pose
is first estimated, typically using discriminatively trained part
models or joint predictors, and then lifted into 3D. While regression
based methods suffer from the need to annotate all images with ground
truth 3D poses -- a technically complex and elaborate process -- for
pipeline approaches the challenge is how to account for uncertainty in
the measurements.  Crucial to both types of approaches is the question
of how to incorporate the 3D dependencies between the different body
joints or to leverage other useful 3D geometric information in the
inference process.


Many earlier works on human pose estimation from a single image
relied on discriminatively trained models to learn a direct mapping
from image features such as silhouettes, HOG or SIFT, to 3D human
poses without passing through 2D landmark estimation
~\cite{agarwal2006recovering,Elgammal:Lee:CVPR:2004,conf/mlmi/EkTL07,Mori:Malik:PAMI:2006,Sigal:etal:CVPR:2009}.

 Recent direct approaches make use of deep
learning~\cite{li20143d, li2015maximum, tompson2014joint,
  toshev2014deeppose}. Regression-based approaches train an end-to-end network 
 to predict 3D joint locations directly from the
image~\cite{toshev2014deeppose, li20143d,
  li2015maximum,zhou2016deep}. Li~\etal~\cite{li2015maximum} incorporate
model joint dependencies in the CNN via a max-margin formalism,
others~\cite{zhou2016deep} impose kinematic constraints by embedding a
differentiable kinematic model into the deep learning architecture.
 Tekin \etal \cite{tekin2016structured} propose a deep
regression architecture for structured prediction that combines
traditional CNNs for supervised learning with an auto-encoder that
implicitly encodes 3D dependencies between body parts.

As CNNs have become more prevalent, 2D joint
estimation~\cite{wei2016convolutional} has become increasingly
reliable and many recent works have looked to exploit this using a
pipeline approach. Papers such as~\cite{chen2014articulated,
  jain2013learning, tompson2014joint, pfister2015flowing} first
estimate 2D landmarks and later 3D spatial relationships are imposed
between them using structured learning or graphical models.

Simo-Serra~\etal~\cite{simo2012single} were one of the first to
propose an approach that naturally copes with the noisy
detections inherent to off-the-shelf body part detectors by modeling
their uncertainty and propagating it through 3D shape space while
satisfying  geometric and kinematic 3D constraints. The work~\cite{sanzari2016bayesian} also estimates the
location of 2D joints before predicting 3D pose using appearance and
the probable 3D pose of discovered parts using a 
non-parametric model.  Another recent example is
Bogo~\etal~\cite{bogo2016keep}, who 
fit a detailed statistical 3D body
model~\cite{loper2015smpl} to 2D joint proposals.

Zhou~\etal~\cite{zhou2015sparseness} tackles the
problem of 3D pose estimation for a monocular image sequence
integrating 2D, 3D and temporal information to account for
uncertainties in the model and the measurements. 
Similar to our proposed approach, Zhou \etal's method~\cite{zhou2015sparseness}  does not need
synchronized 2D-3D training data, \ie it only needs 2D pose
annotations to train the CNN joint regressor and a separate 3D mocap
dataset to learn the 3D sparse basis.  Unlike our approach, it relies
on temporal smoothness for its best performance, and performs poorly on 
a single image.

Finally, Wu~\etal~\cite{wu2016single}'s 3D Interpreter Network, a
recent approach to estimate the skeletal structure of common objects
(chairs, sofas, ...) bears similarities with our method. Although our
approaches share common ground in the decoupling of 3D and 2D training
data and the use of projection from 3D to improve 2D predictions the
network architectures are very different and, unlike us, they do not
carry out a quantitative evaluation on 3D human pose
estimation.

\section{Network Architecture}

Figure~\ref{fig:pipeline} illustrates the main contribution of our
approach, a new multi-stage CNN architecture that can be trained
end-to-end to estimate jointly 2D and 3D joint locations. Crucially it
includes a novel layer, based on a probabilistic 3D model of human
pose, responsible for lifting 2D poses into 3D and propagating 3D
information about the skeletal structure to the 2D convolutional
layers. In this way, the prediction of 2D pose benefits from the 3D
information encoded. Section~\ref{sec:modeling-3D-poses} describes the
new probabilistic 3D model of human pose, trained on a dataset of 3D
mocap data. Section~\ref{sec:pose-inference} describes all the new
components and layers of the  CNN architecture. Finally,
Section~\ref{sec:exper-eval} describes experimental evaluation on
the Human3.6M dataset where we obtain state-of-the-art results. In
addition we show qualitative results on images from the MPII and Leeds datasets.

\section{Probabilistic 3D Model of Human Pose}
\label{sec:modeling-3D-poses}

One fundamental challenge in creating models of human poses lies in
the lack of access to 3D data of sufficient variety to characterize
the space of human poses. To compensate for this lack of data
we identify and eliminate confounding factors such as rotation in the
ground plane, limb length, and left-right symmetry that lead to
conceptually similar poses being unrecognized in the training data.

Simple preprocessing eliminates some factors. Size variance is addressed by
normalizing the data such that the sum of squared limb lengths on the human
skeleton is one; while left-right symmetry is exploited by flipping each pose in
the x-axis and re-annotating left as right and vice-versa.

\subsection{Aligning 3D Human Poses in the Training Set}
\label{sec:aligning-non-rigid}

Allowing for rotational invariance in the ground-plane is more
challenging and requires integration with the  data model. We seek the
optimal rotations for each pose such that after rotating the poses
they are closely approximated by a low-rank compact Gaussian
distribution.

We formulate this as a problem of optimization over a set of
variables. Given a set of $N$ training 3D poses, each represented as a
$(3 \times L)$ matrix ${\bf P_i}$ of 3D landmark locations, where
$i\in\{1,2,..,N\}$ and $L$ is the number of human joints/landmarks; we
seek global estimates of an average 3D pose $\mu$, a set of $J$
orthonormal basis matrices\footnote{When we say ${\bf e}$ is a set of
  orthonormal basis matrices we mean that each matrix, if unwrapped
  into a vector, is of unit norm and orthogonal to all other unwrapped
  matrices.} ${\bf e}$ and noise variance $\sigma$, alongside per
sample rotations $R_i$ and basis coefficients $a_i$ to minimize
the following estimate 
\begin{align}
\argmin_{{\bf R}, \mu , a, {\bf e}, \sigma} \sum_{i=1}^{N} \big( & ||{\bf P_i} - {\bf R_i}\left({\bf \mu} +a_i \cdot {\bf e}\right)||_2^2 \\
+                                                                   & \sum_{j=1}^{J} (a_{i,j}\cdot\sigma_j)^2
 +\ln \sum_{j=1}^{J} \sigma_j^2 \big)\nonumber
\end{align}

Where $a_i\cdot{\bf e}= \sum_j a_{i,j} {\bf e_j}$ is the tensor analog
of a multiplication between a vector and a matrix, and $||\cdot||_2^2$ is
the squared Frobenius norm of the matrix. Here the $y$-axis is assumed to point up, and the rotation matrices
$R_i$ considered are ground plane rotations. 
With the large number of 3D pose samples considered (of the order of 1 million
when training on the Human3.6M dataset~\cite{ionescu2014human3}), and the
complex inter-dependencies between samples for ${\bf e}$ and $\sigma$, the
memory requirements mean that it is not possible to solve directly as a joint
optimization over all variables using a non-linear solver such as Ceres.
Instead, we carefully initialize and alternate between performing closed-form
PPCA~\cite{Tipping99probabilisticprincipal} to update $\mu , a, {\bf e},
\sigma$; and updating ${R_i}$ using Ceres~\cite{ceres-solver} to minimize the
above error.
%
As we do this, we steadily increase the size of the basis from $1$
through to its target size $J$.  This stops apparent deformations that
could be resolved through rotations from becoming locked into the
basis at an early stage, and empirically leads to lower cost
solutions.

To initialize we use a variant of the Tomasi-Kanade~\cite{Tomasi1992}
algorithm to estimate the mean 3D pose ${\mu}$. As the $y$
component is not altered by planar rotations, we take as our estimate
of the $y$ component of ${\mu}$, the mean of each point in the
$y$ direction. For the $x$ and $z$ components, we interleave the $x$
and $z$ components of each sample and concatenate them into a large
$2\,N \times L$ matrix ${\bf M}$, and find the
rank two approximation of this such that ${\bf M} \approx {\bf A}\cdot {\bf B}$. We
then calculate $ {\bf \hat A}$ by replacing each adjacent pair of rows
of ${\bf A}$ with the closest orthonormal matrix of rank two, and take
${\bf \hat A^{\dag} M}$ as our estimate\footnote{${\bf A}^\dag$ being the
  pseudo-inverse of ${\bf A}$.}  of the $x$ and $z$
components of $\mu$.

\begin{figure}[tb]
  \begin{center}
    \vspace{-7mm}
  \includegraphics[width=0.9\linewidth]{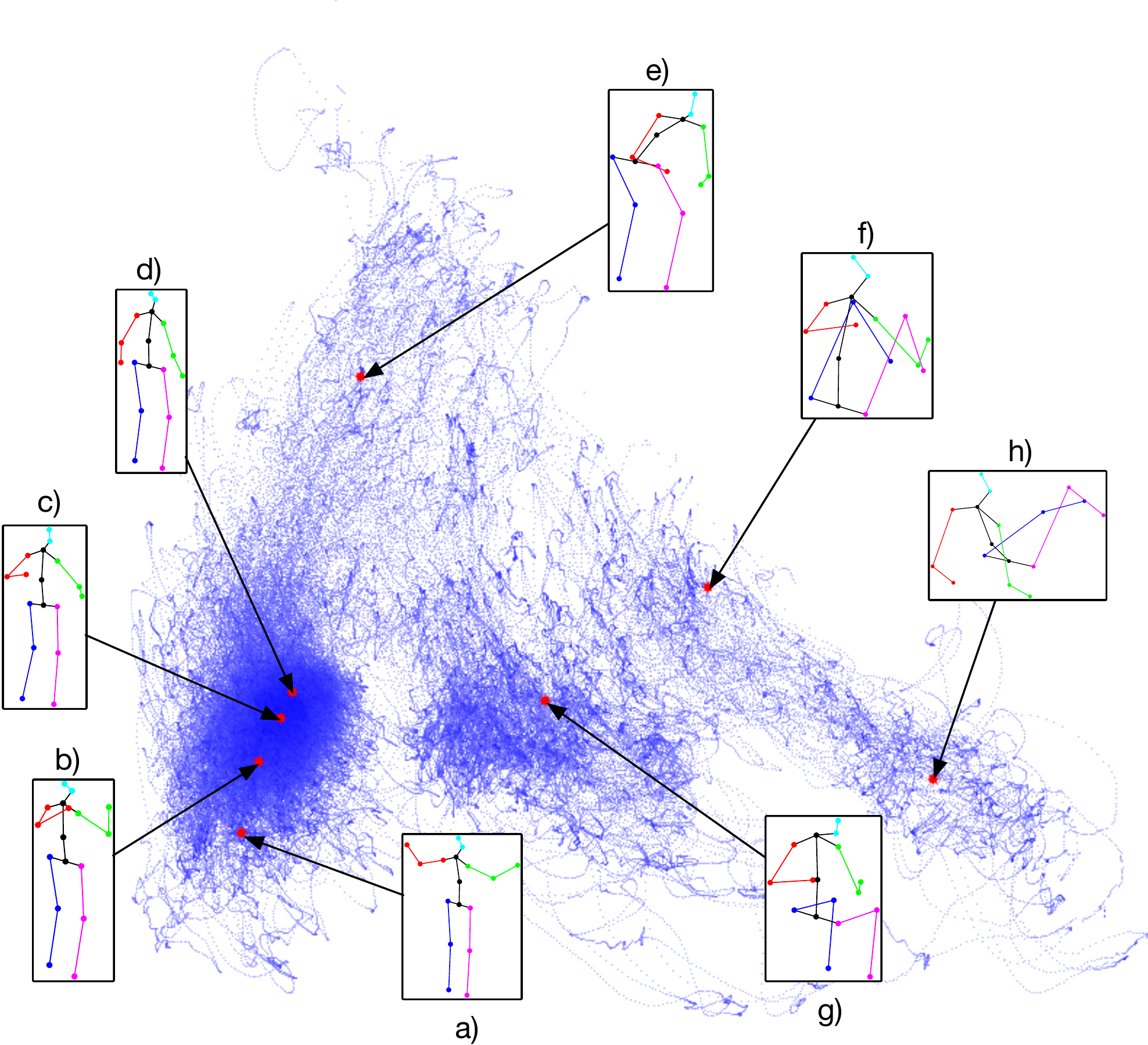}
\end{center}
    \vspace{-7mm}
   \caption{\small Visualization of the 3D training data after alignment (see
     section~\ref{sec:aligning-non-rigid}) using 2D PCA. Notice how
     all poses have the same orientation.  \emph{Standing-up} poses
     a), b), c) and d) are all close to each other and far from
     \emph{sitting-down} poses f) and h) which form another clear
     cluster.
     \label{fig:manifold}\vspace{-7mm}}
\end{figure}

The end result of this optimization is a compact low-rank
approximation of the data in which all reconstructed poses appear to
have the same orientation (see Figure \ref{fig:manifold}). In the next
section we extend the  model to be described as a multi-modal
distribution to better capture the variations in the space of 3D human
poses.
\subsection{A Multi-Modal Model of 3D Human Pose}
\label{sec:learn-align}
Although the learned Gaussian model of
section~\ref{sec:aligning-non-rigid} can be directly used to estimate the 3D
(see Table \ref{tab:comparison_1}), inspection of figure
\ref{fig:manifold} shows that the data is not Gaussian distributed and
is better described using a multiple modal distribution. In doing
this, we are heavily inspired both by approaches such
as~\cite{pitelis2013learning} which characterize the space of human
poses as a mixture of PCA bases, and by related works such
as~\cite{wang2014representing,bregler2000recovering} that represent
poses as an interpolation between exemplars. These approaches are
extremely good at modeling tightly distributed poses (e.g. walking)
where samples in the testing data are likely to be close to poses seen
in training.  This is emphatically not the case in much of the
Human3.6M dataset, which we use for evaluation. Zooming in on the
edges of Figure~\ref{fig:manifold} reveals many isolated paths where
motions occur once and are never revisited again.

Nonetheless, it is precisely these regions of low-density that we are
interested in modeling. As such, we seek a coarse representation of
the pose space that says something about the regions of low density
but also characterizes the multi-modal nature of the pose space.  We
 represent the  data as a mixture of probabilistic PCA models
using few clusters, and trained using the
EM-algorithm~\cite{Tipping99probabilisticprincipal}. When using a
small number of clusters, it is important to initialize the algorithm
correctly, as accidentally initializing with multiple clusters about a
single mode, can lead to poor density estimates. To initialize we make use of a simple heuristic.

We first subsample the aligned poses (which we refer to as $P$), and
then compute the Euclidean distance $d$ among pairs.  We seek a set
of $k$ samples $S$ such that the distance between points and their
nearest sample is minimized
\begin{equation}
\argmin_S \sum_{p\in P} \min_{s\in S} d(s,p)
\label{eq:select}
\end{equation}
We find $S$ using greedy selection, holding our previous estimate of
$S$ constant, and iteratively selecting the next candidate $s$ such
that $\{s\}\cup S$ minimizes the above cost. A selection of 3D pose
samples found using this procedure can be seen in the rendered poses
of Figure~\ref{fig:manifold}. In practice, we stop proposing
candidates when they occur too close to the existing candidates, as
shown by samples (a--d), and only choose one candidate from the
dominant mode.

Given these candidates for cluster centers, we assign each aligned
point to a cluster representing its nearest candidate and then run the
EM algorithm of~\cite{Tipping99probabilisticprincipal}, building a mixture of probabilistic PCA bases.
\begin{figure*}\vspace{-5mm}
\hspace{-9mm}
\begin{tabular}{p{1.2\columnwidth}p{1mm}p{0.8\columnwidth}}
\begin{center}
\setlength{\tabcolsep}{1pt}
\begin{tabular}{cccccc}
\small Stage 1                                                     & \small Stage 2      & \small Stage 3      & \small Stage 4      & \small Stage 5      & \small Stage 6      \\
\includegraphics[width=0.16\linewidth]{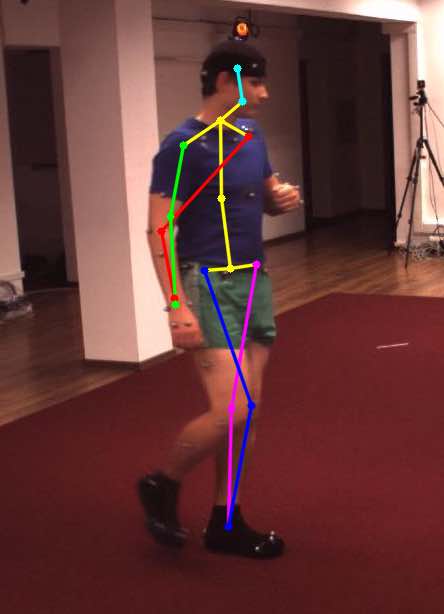}    & 
\includegraphics[width=0.16\linewidth]{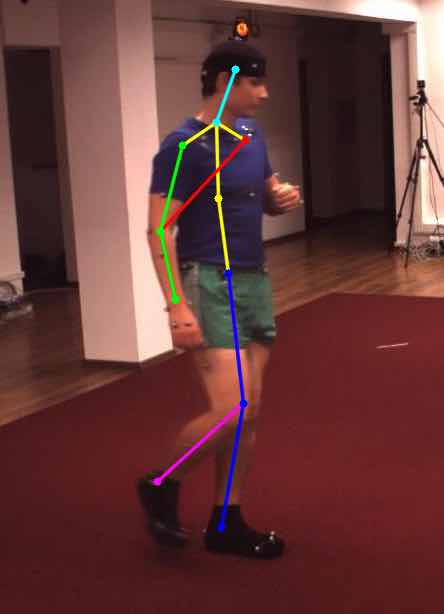}    & 
\includegraphics[width=0.16\linewidth]{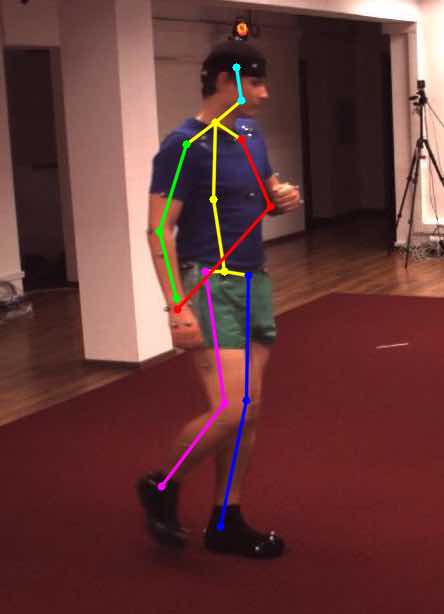}    & 
\includegraphics[width=0.16\linewidth]{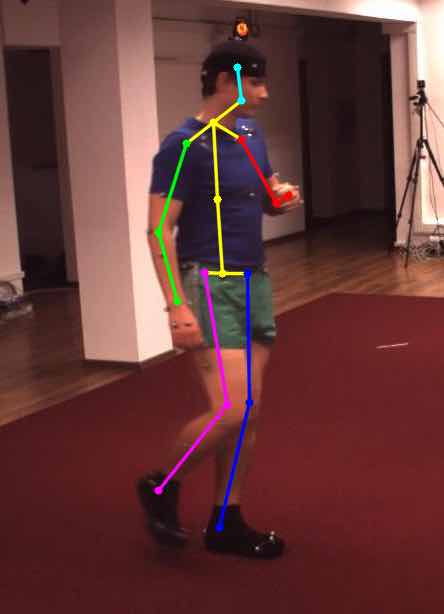}    & 
\includegraphics[width=0.16\linewidth]{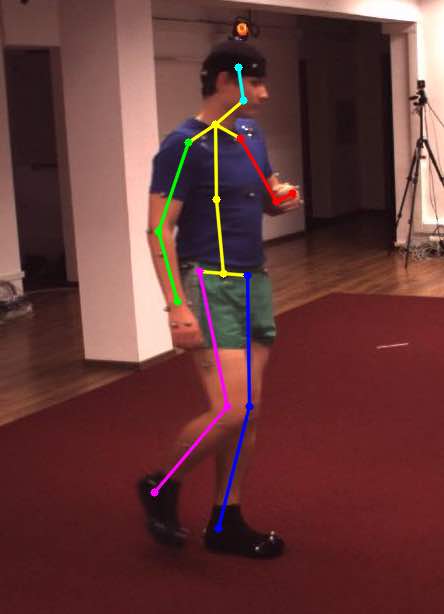}    & 
\includegraphics[width=0.16\linewidth]{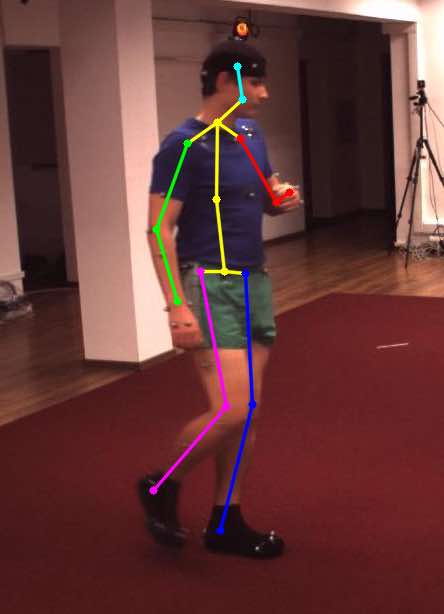}    \\
\includegraphics[width=0.16\linewidth]{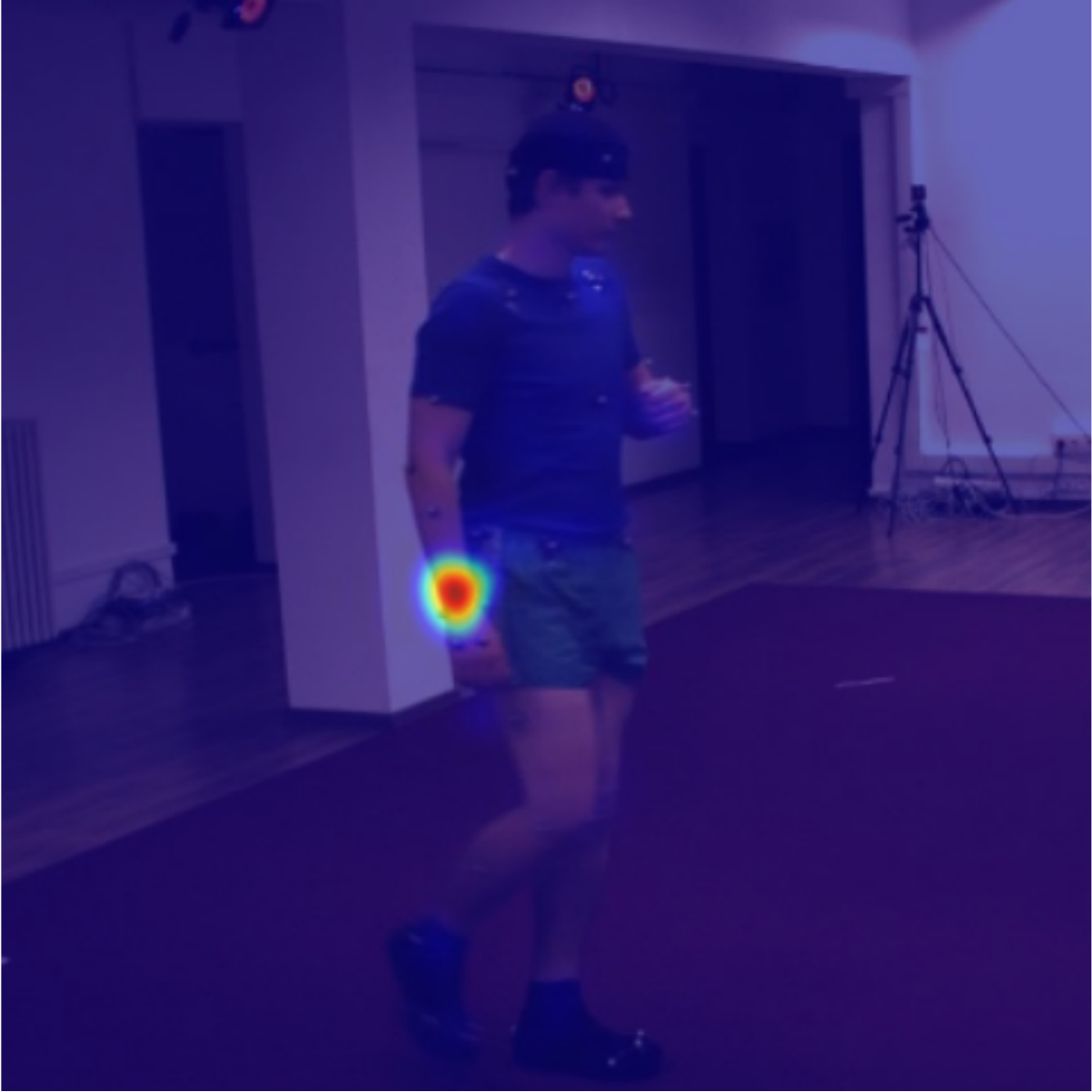} & 
\includegraphics[width=0.16\linewidth]{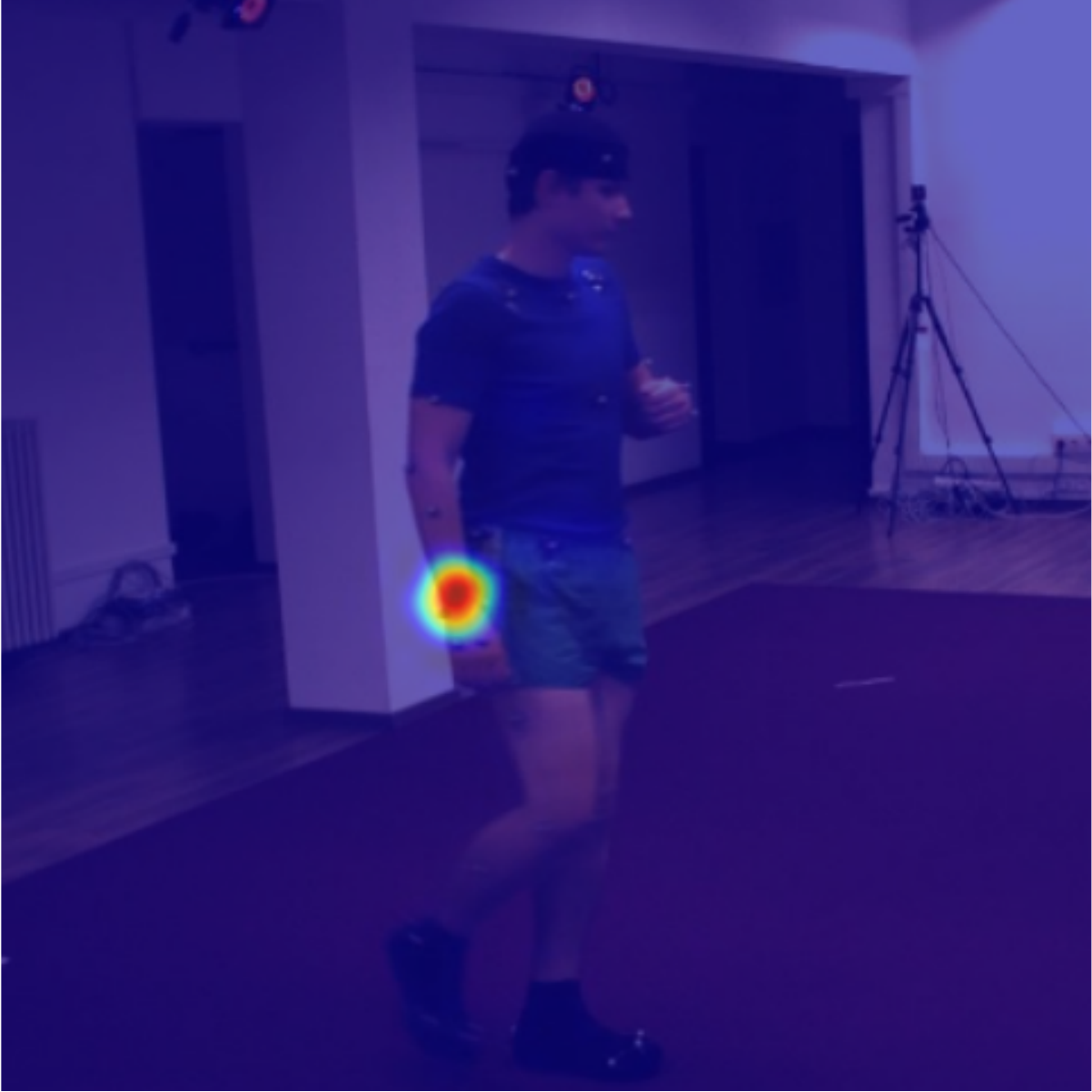} & 
\includegraphics[width=0.16\linewidth]{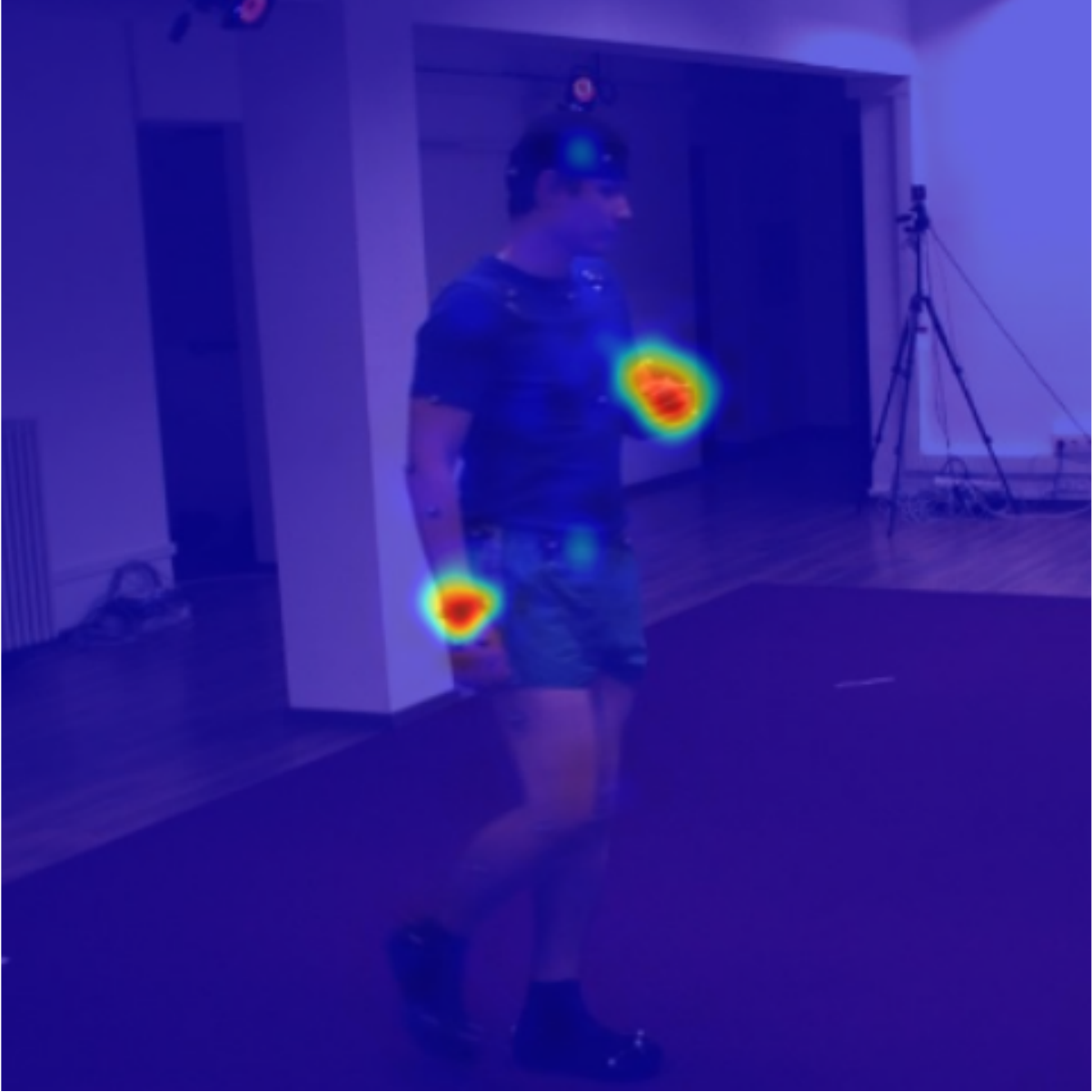} & 
\includegraphics[width=0.16\linewidth]{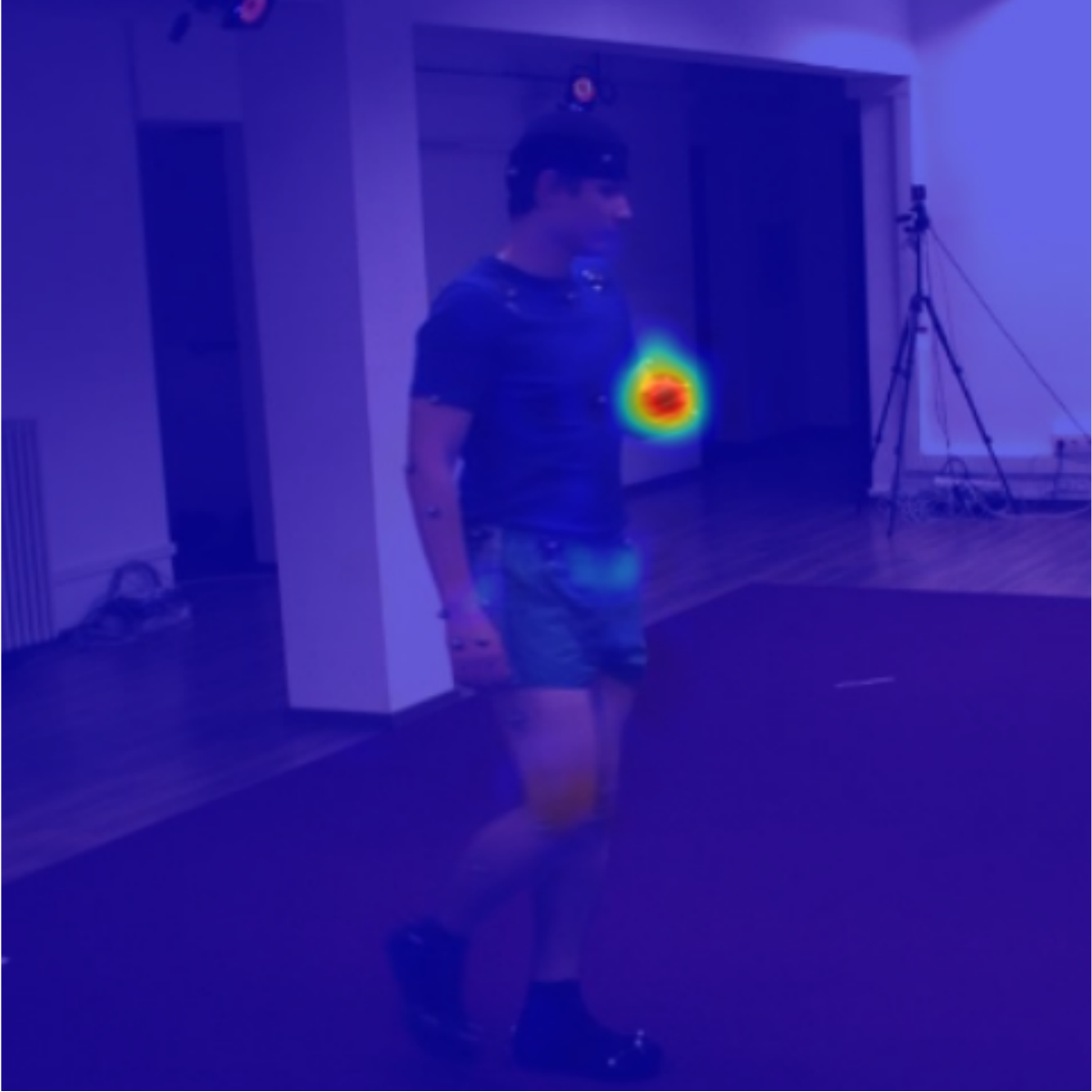} & 
\includegraphics[width=0.16\linewidth]{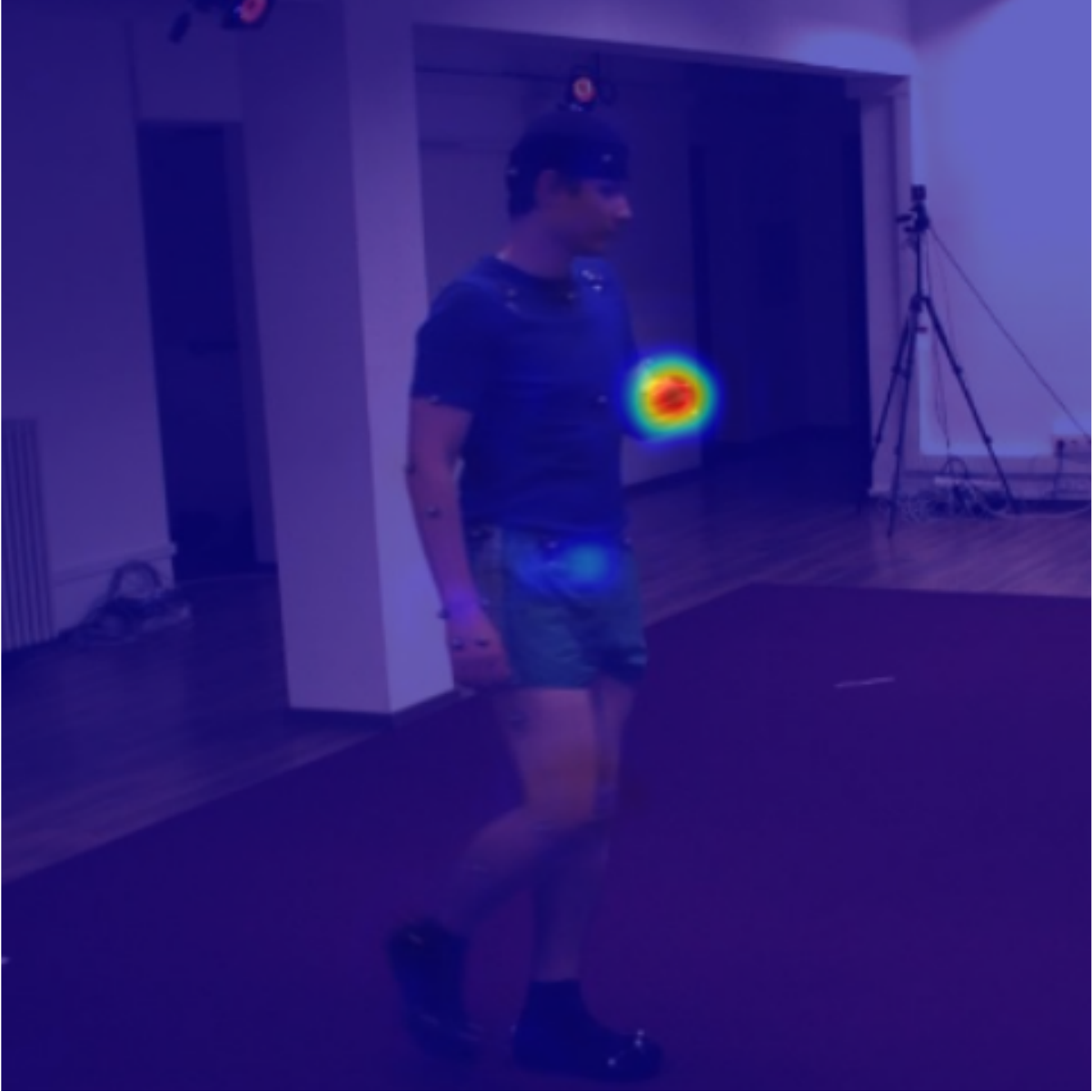} & 
\includegraphics[width=0.16\linewidth]{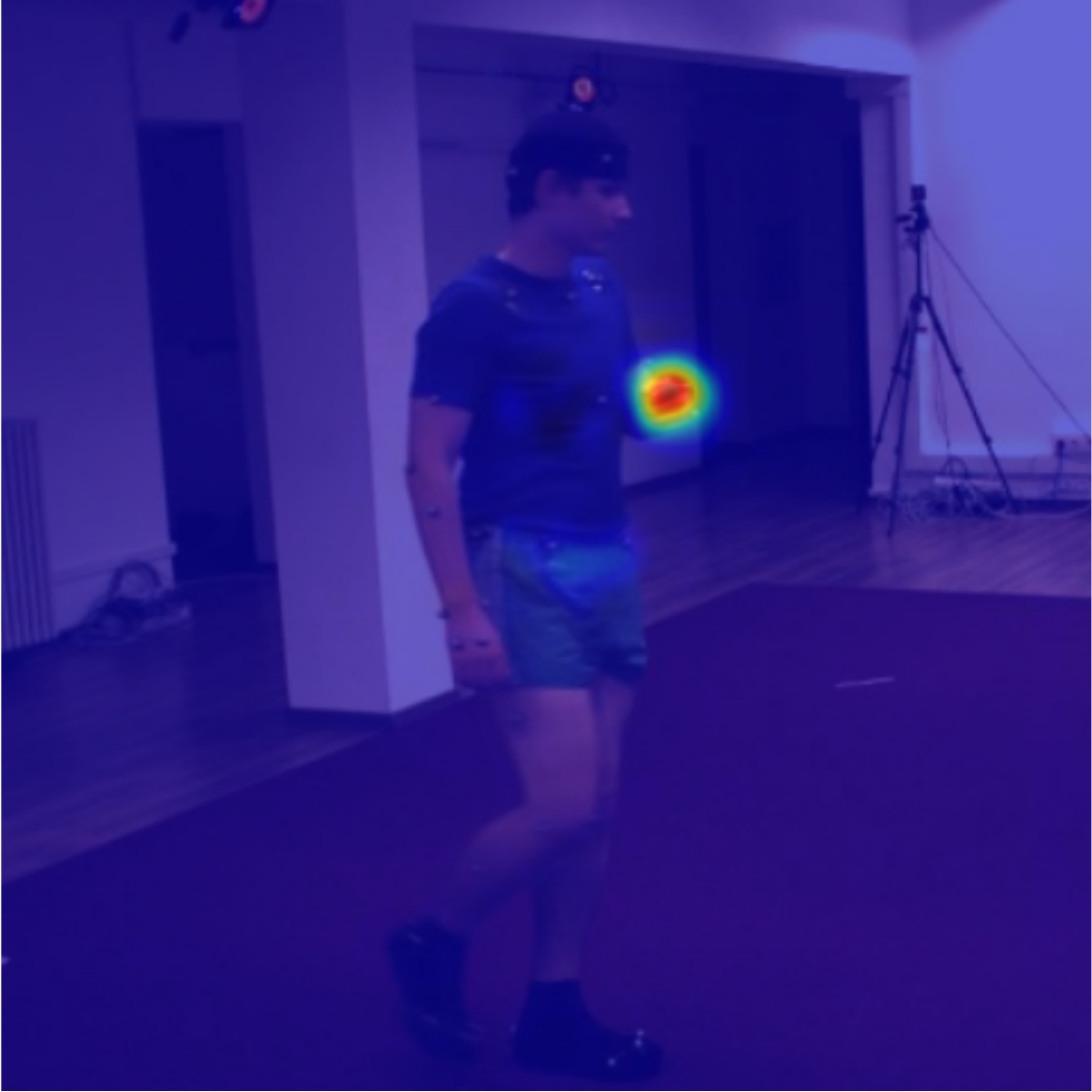}                        \\
\end{tabular}
\end{center}
                                                                   &&
\begin{center}
\vspace{5mm}
  \setlength{\tabcolsep}{2pt}
  \begin{tabular}{cccccc}
\small Stage 1                                                     & \small Stage 2      & \small Stage 3      & \small Stage 4      & \small Stage 5      & \small Stage 6      \\
\includegraphics[height=0.4\linewidth]{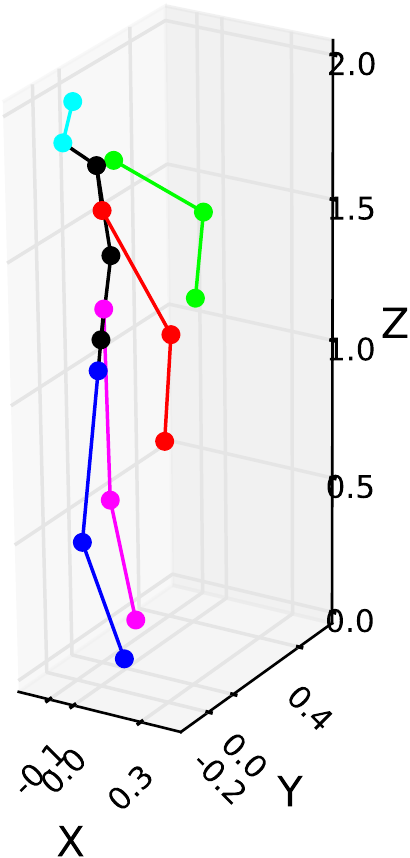}         & 
\includegraphics[height=0.4\linewidth]{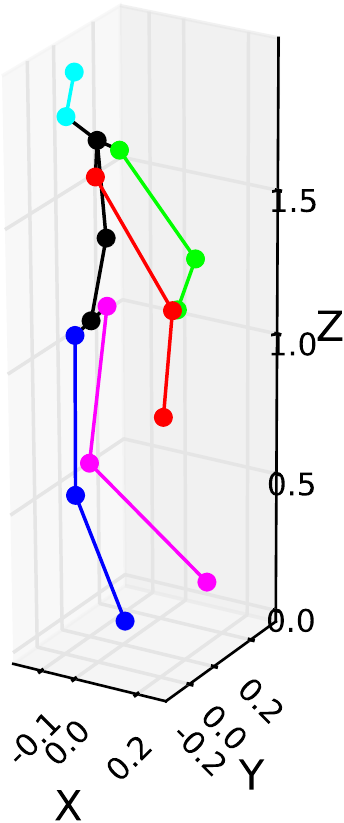}         & 
\includegraphics[height=0.4\linewidth]{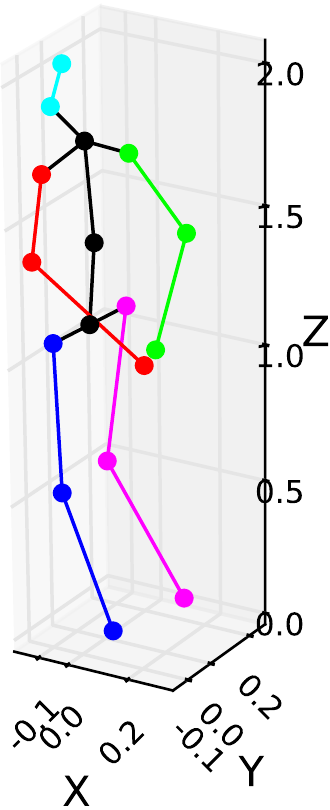}         & 
\includegraphics[height=0.4\linewidth]{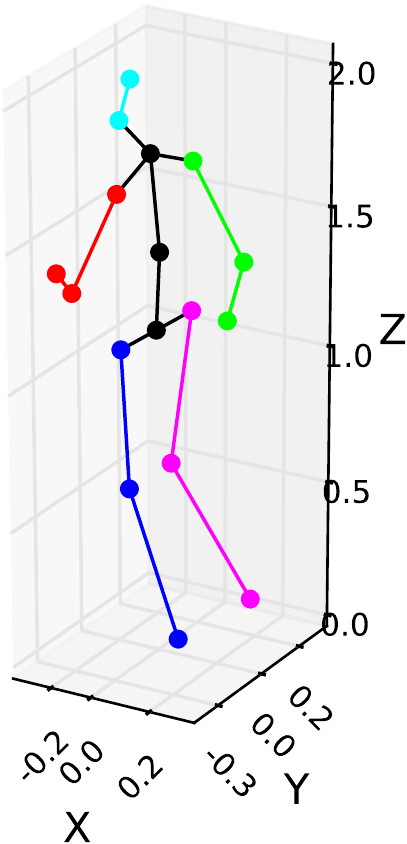}         & 
\includegraphics[height=0.4\linewidth]{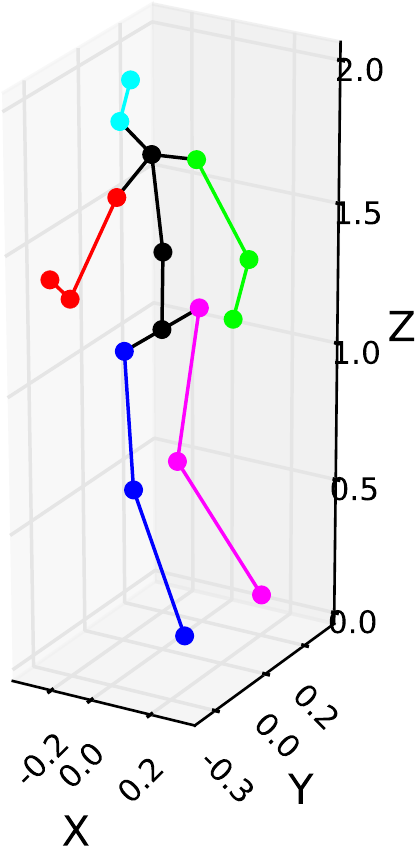}         & 
\includegraphics[height=0.4\linewidth]{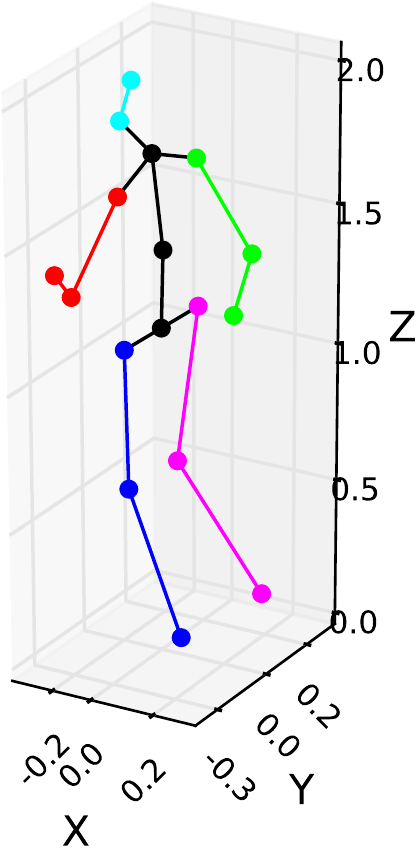}                                                                                                                       \\
  \small  9.19 mm                                                & \small  7.30 mm & \small  6.64 mm & \small  3.34 mm & \small  3.28 mm & \small  3.10 mm
\end{tabular}
\end{center}
\end{tabular}
\vspace{-5mm}
\caption{\small Results returned by different stages of the
  architecture. \textit{Top Left}:~Evolution of the 2D skeleton after
  projecting the 3D points back into the 2D space;
  \textit{Bottom Left}:~Evolution of the beliefs for the landmark {\em Left
  hand} through the stages. \textit{Right}:~3D skeleton
  with the relative mean error per landmark in
  millimeters. Even with incorrect
  landmark locations, the model returns a physically
  plausible solution.\label{fig:manifold_3D_corrections_stages}\vspace{-5mm}}
\end{figure*}

\section{A New Convolutional Architecture for 2D and 3D Pose Inference}
\label{sec:pose-inference}
Our 3D pose inference from a single RGB image makes use of a
multistage deep convolutional architecture, trained end-to-end, that
repeatedly fuses and refines 2D and 3D poses, and a second module
which takes the final predicted 2D landmarks and lifts them one last
time into 3D space for the  final estimate (see
Figure~\ref{fig:pipeline}).

At its heart, the  architecture is a novel refinement of the
Convolutional Pose Machine of Wei~\etal~\cite{wei2016convolutional}, who reasoned exclusively in
2D, and proposed an architecture that iteratively refined 2D pose
estimations of landmarks using a mixture of knowledge of the image and
of the estimates of landmark locations of the previous stage. We
modify this architecture by generating, at each stage, projected 3D
pose belief maps which are fused in a learned manner with the standard
maps. From an implementation point of view this is done by introducing
two distinct layers, the \textit{probabilistic 3D pose layer} and the
\textit{fusion layer} (see Figure~\ref{fig:pipeline}).

Figure~\ref{fig:manifold_3D_corrections_stages} shows how the 2D
uncertainty in the belief maps is reduced at each stage of the
architecture and how the accuracy of the 3D poses increases with each
stage. 

\subsection{Architecture of each stage}
The sequential architecture consists of 6 stages. Each stage consists of 4
distinct components (see Figure~\ref{fig:pipeline}):

\noindent{\bf Predicting CNN-based belief-maps:} we use a set of
convolutional and pooling layers, equivalent to those used in the
original CPM architecture~\cite{wei2016convolutional}, that combine
evidence obtained from image learned features with the belief maps
obtained from the previous stage ($t-1$) to predict an updated set of
belief maps for the 2D human joint positions.

\noindent{\bf Lifting 2D belief-maps into 3D:} the output of the
CNN-based belief maps is taken as input to a new layer that uses new
pretrained \emph{probabilistic 3D human pose model} to lift the
proposed 2D poses into 3D. 

\noindent{\bf Projected 2D pose belief maps:} The 3D pose estimated by
the previous layer is projected back onto the image plane to produce a
new set of projected pose belief maps. These maps encapsulate 3D
dependencies between the body parts. 

\noindent{\bf 2D Fusion layer:} The final layer in each stage (described in
section~\ref{sec:fusion_layer}) learns the weights to fuse the two sets of
belief maps into a single estimate passed to the next stage.

\noindent{\bf Final lifting:} The belief maps produced as the output of the
final stage ($t=6$) are then lifted into 3D to give the final estimate for the
pose (see Figure~\ref{fig:pipeline}) using our algorithm to lift 2D poses into
3D.

\begin{figure*}[tbp]\vspace{-5mm}
  \hspace{-13mm}
  \setlength{\tabcolsep}{-3pt}
  \begin{tabular}{p{1.2\columnwidth}p{0.1mm}p{0.8\columnwidth}}
    \begin{center}
      \setlength{\tabcolsep}{1pt}
  \begin{tabular}{cccccc}
    \multicolumn{6}{c}{Results from the Human3.6M dataset}\\
    \toprule
\includegraphics[height=0.14\linewidth]{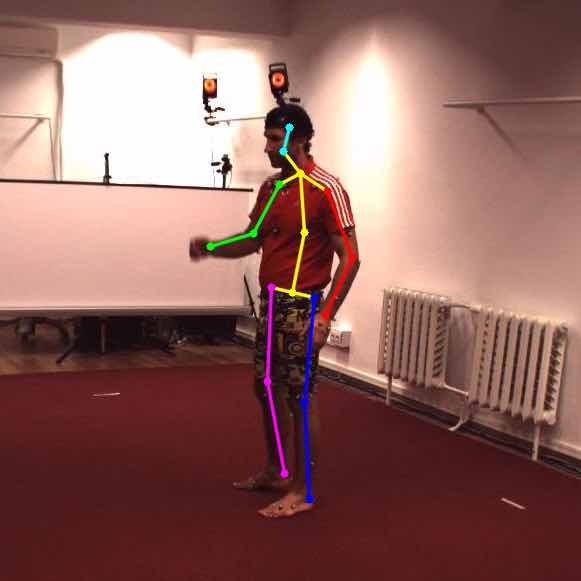}    & 
\includegraphics[height=0.14\linewidth]{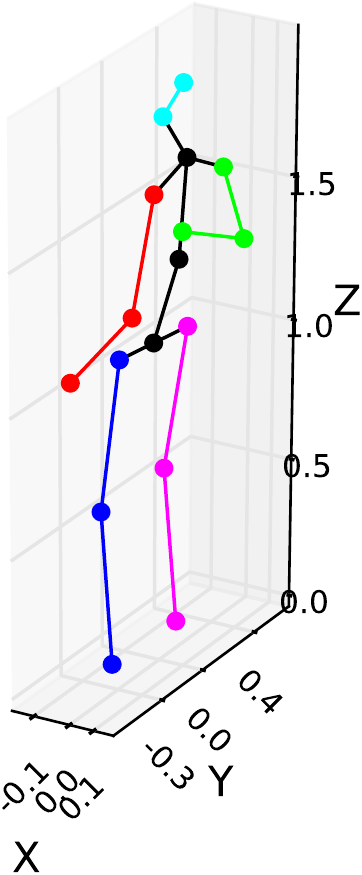}    & 
\includegraphics[height=0.14\linewidth]{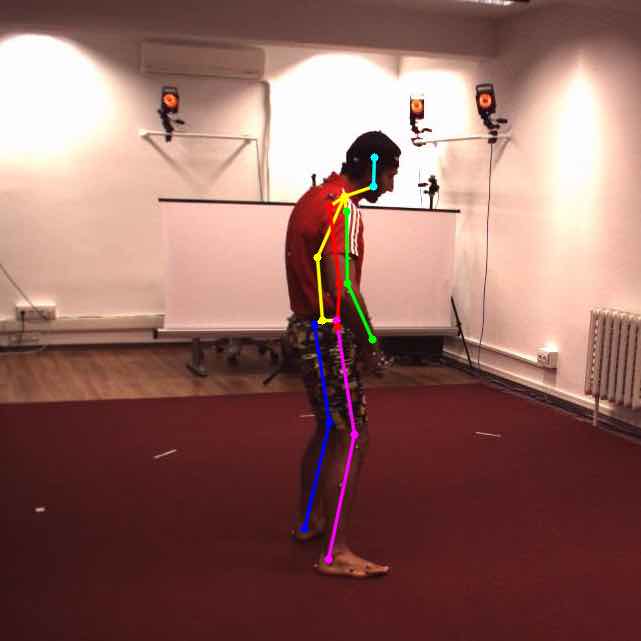}    & 
\includegraphics[height=0.14\linewidth]{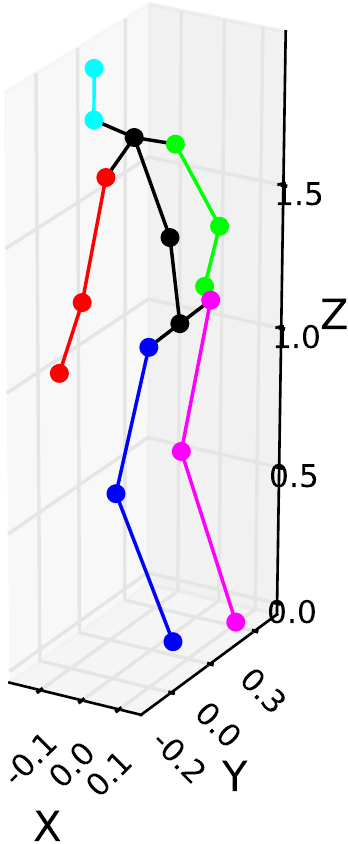}    & 
\includegraphics[height=0.14\linewidth]{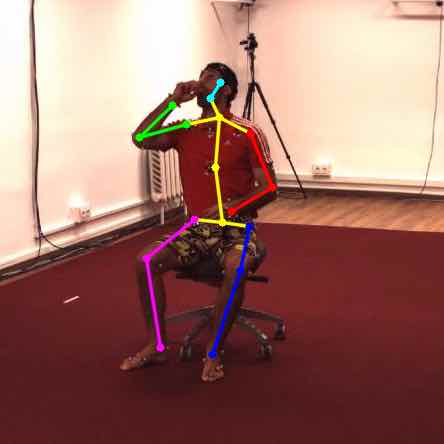}    & 
\includegraphics[height=0.14\linewidth]{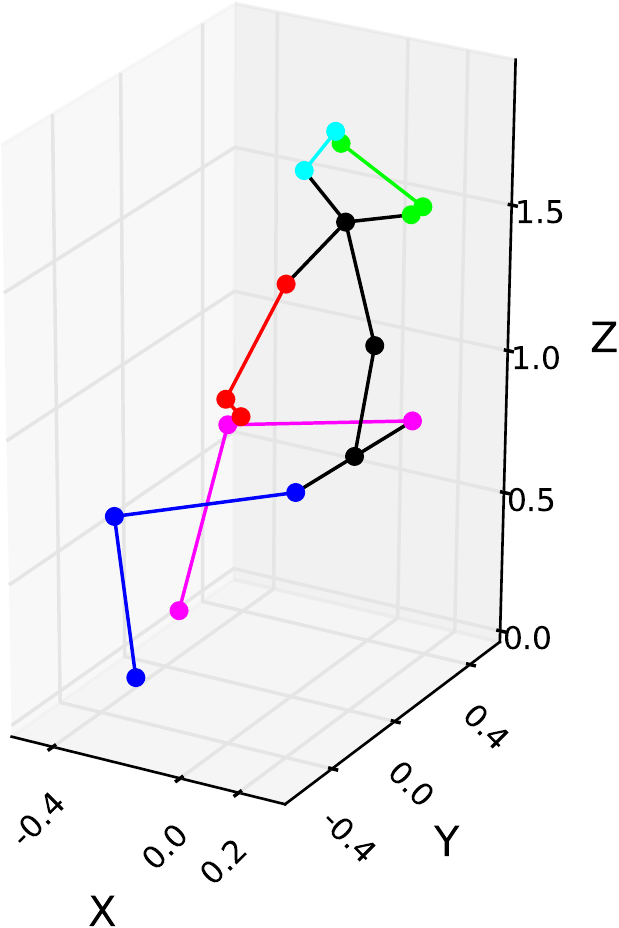}    \\
\includegraphics[height=0.14\linewidth]{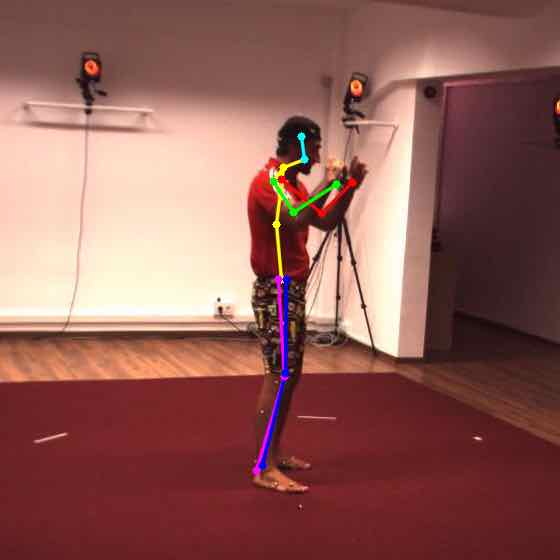}    & 
\includegraphics[height=0.14\linewidth]{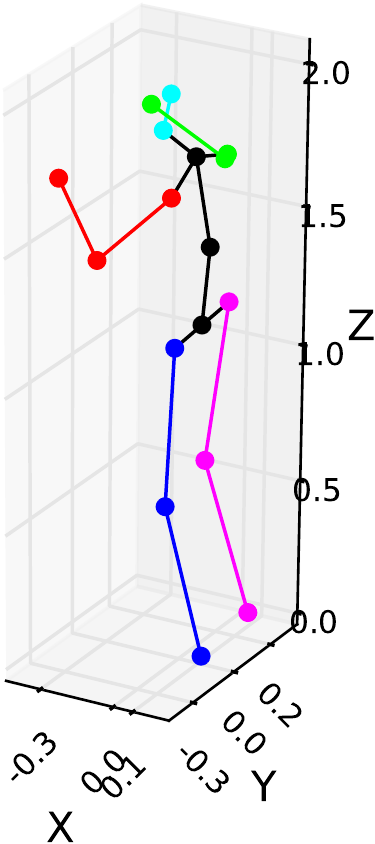}    & 
\includegraphics[height=0.14\linewidth]{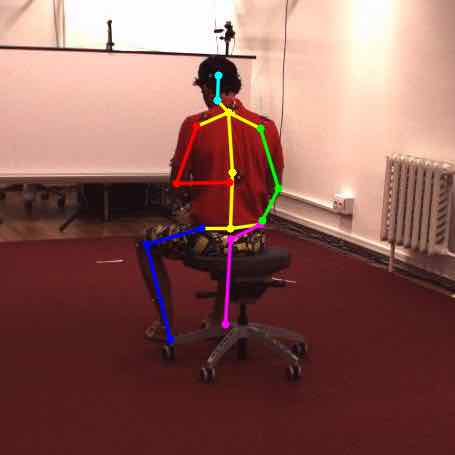}    & 
\includegraphics[height=0.14\linewidth]{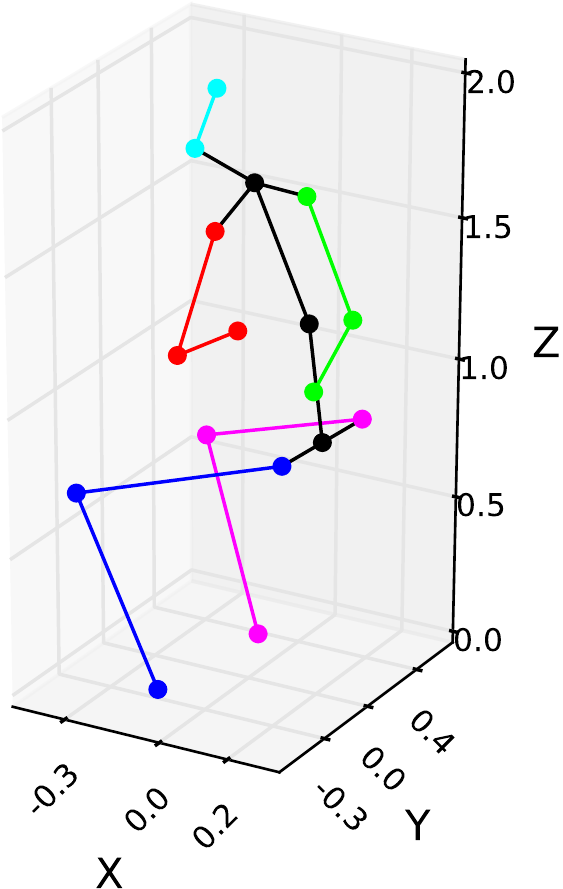}    & 
\includegraphics[height=0.14\linewidth]{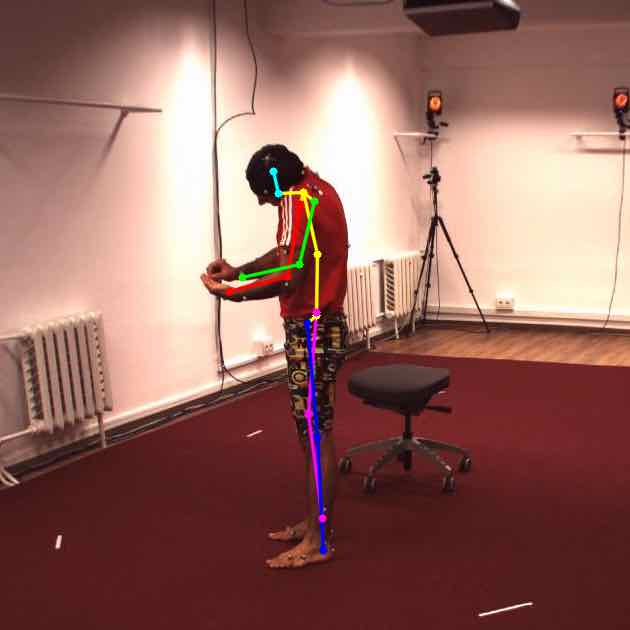}    & 
\includegraphics[height=0.14\linewidth]{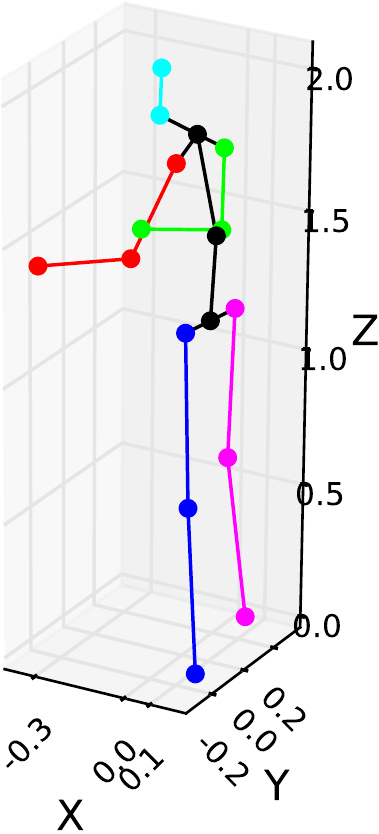}    \\
\includegraphics[height=0.14\linewidth]{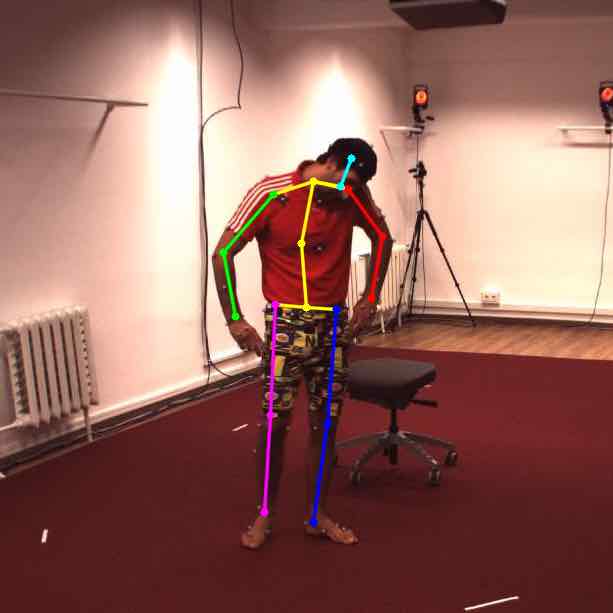}    & 
\includegraphics[height=0.14\linewidth]{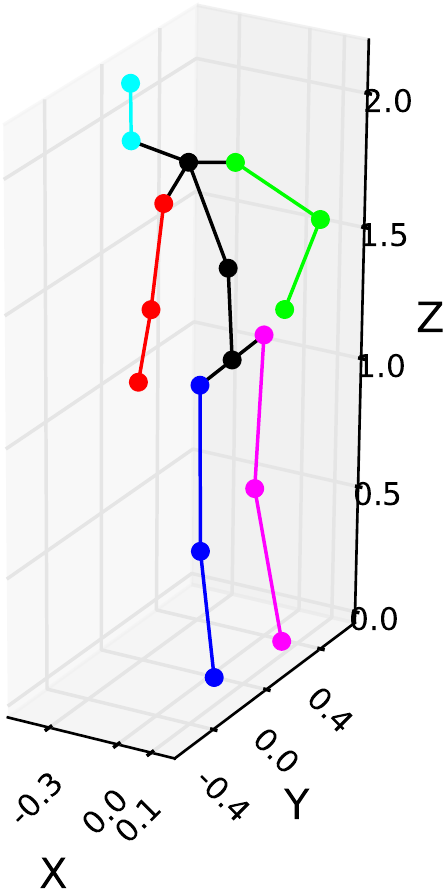}    & 
\includegraphics[height=0.14\linewidth]{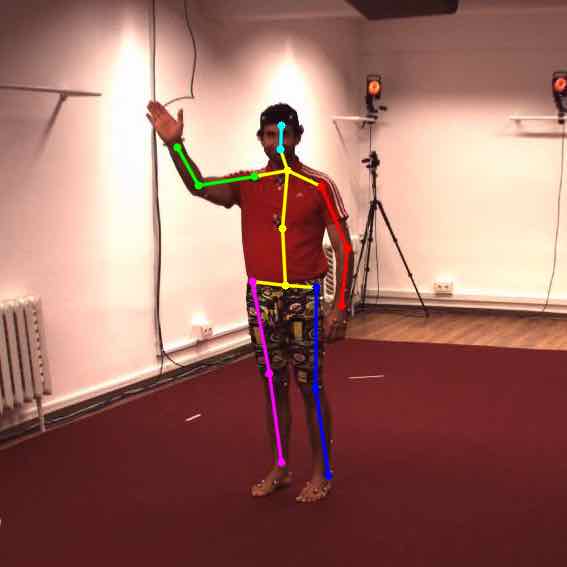}    & 
\includegraphics[height=0.14\linewidth]{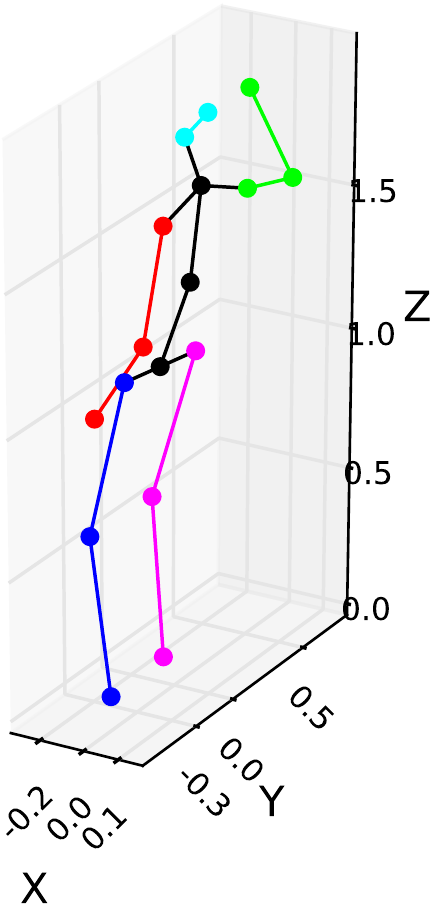}    & 
\includegraphics[height=0.14\linewidth]{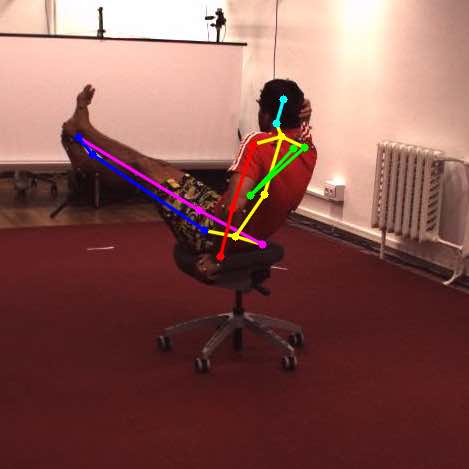}    & 
\includegraphics[height=0.14\linewidth]{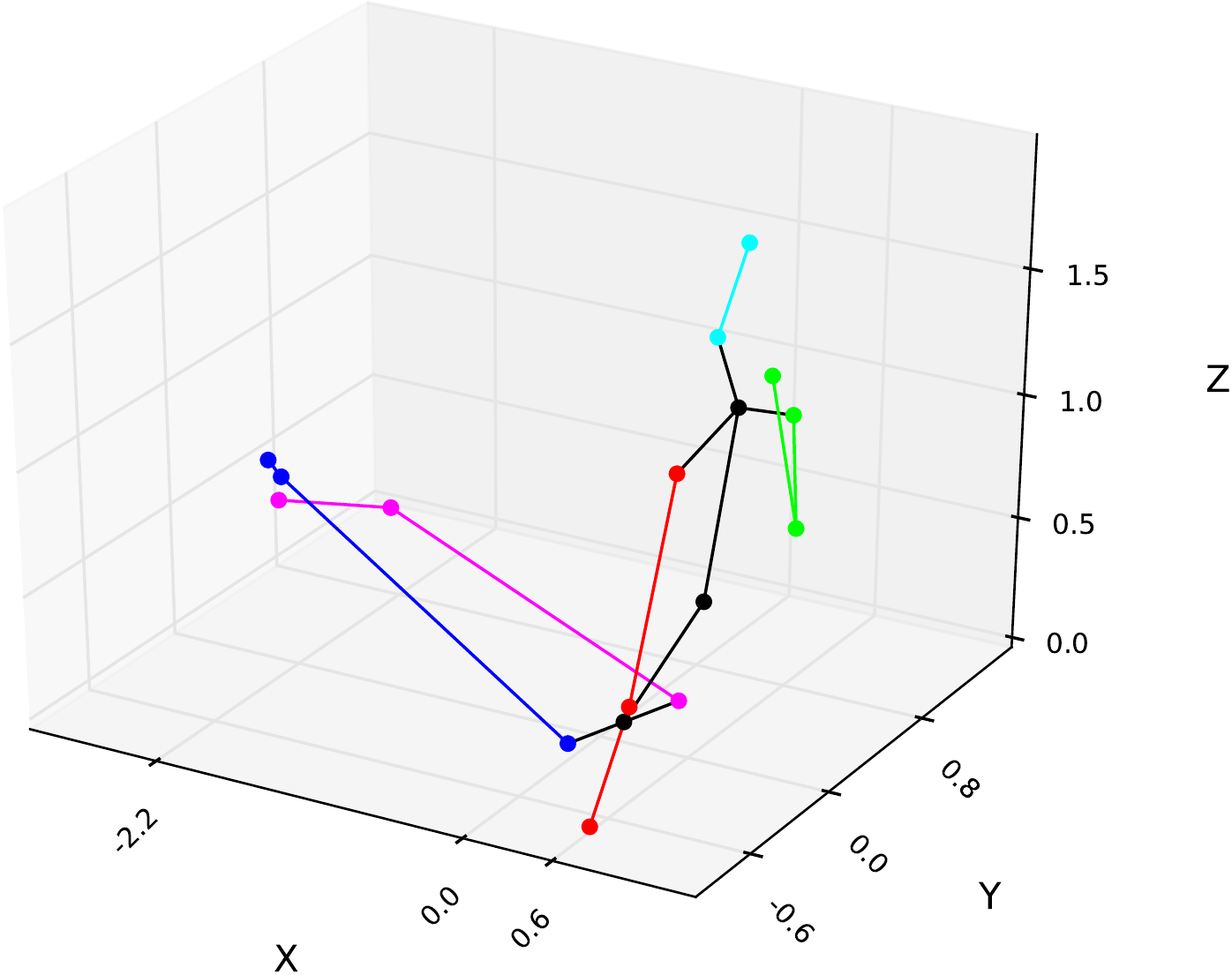}    \\
\includegraphics[height=0.14\linewidth]{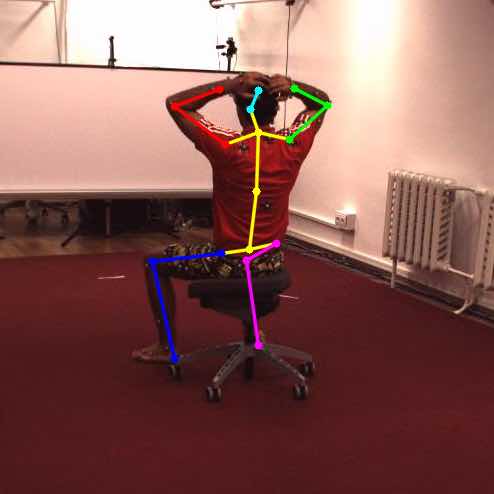}   & 
\includegraphics[height=0.14\linewidth]{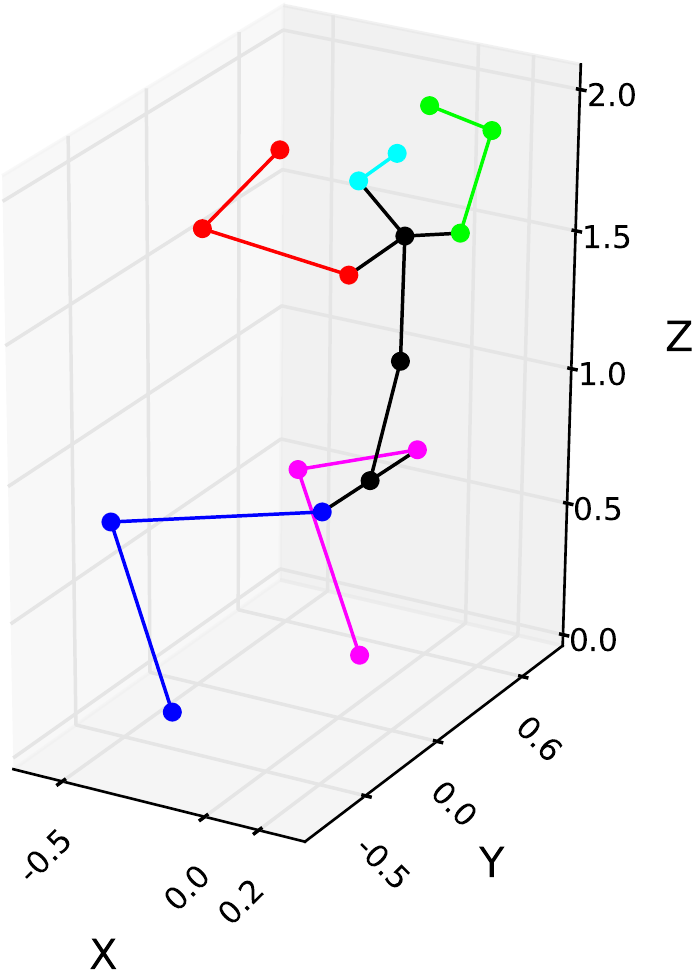}   & 
\includegraphics[height=0.14\linewidth]{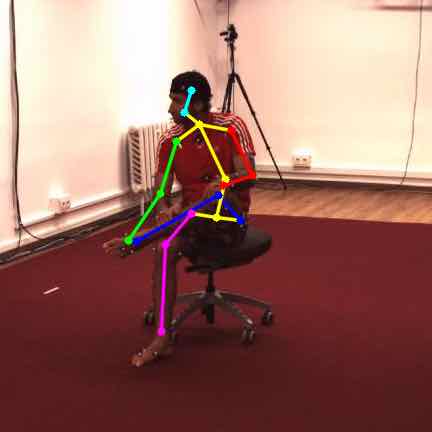}   & 
\includegraphics[height=0.14\linewidth]{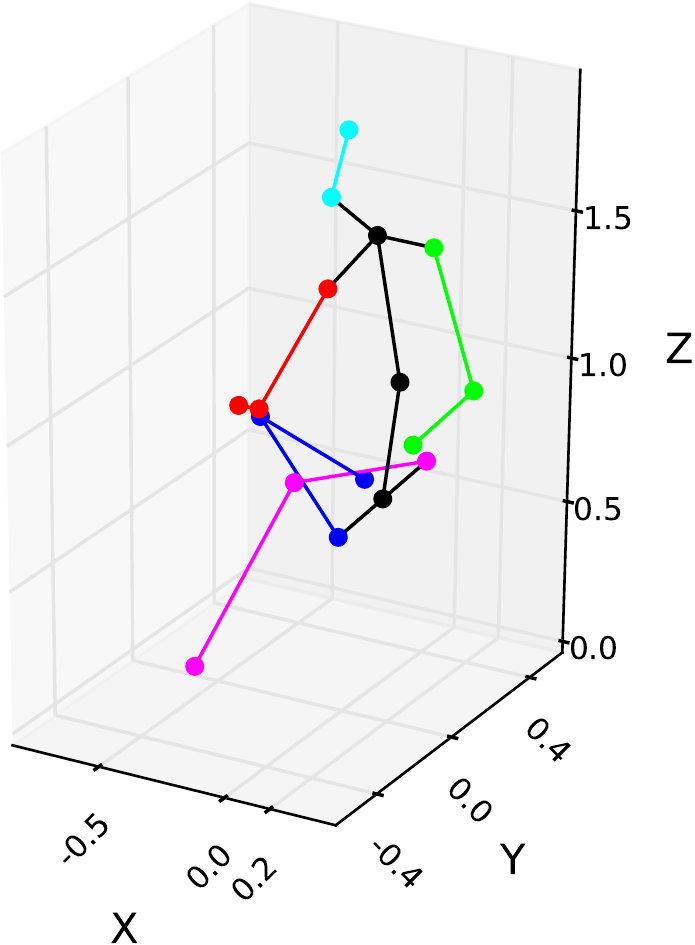}   & 
\includegraphics[height=0.14\linewidth]{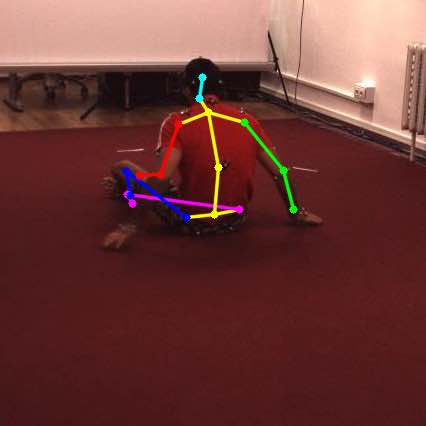}   & 
\includegraphics[height=0.14\linewidth]{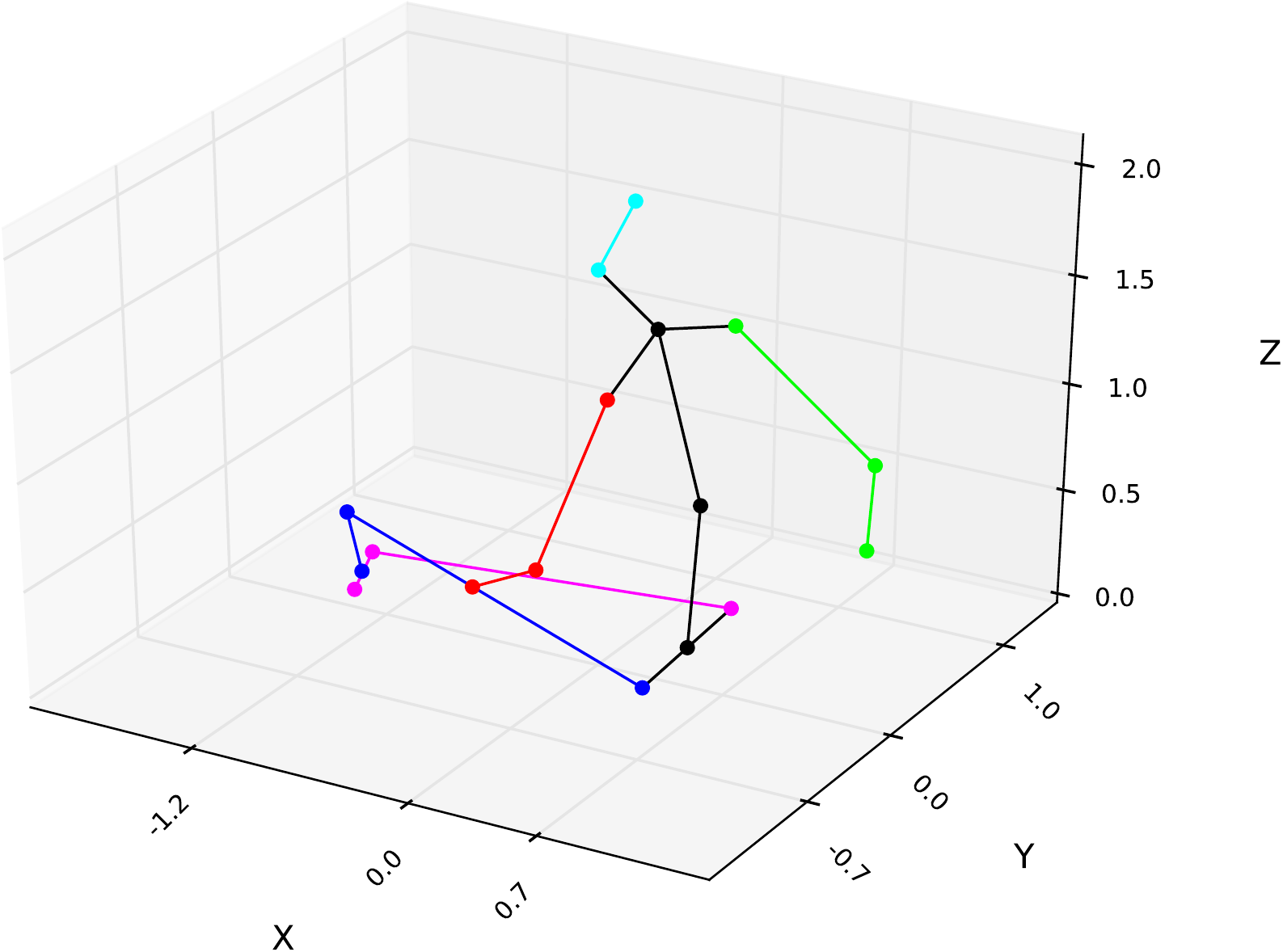}   \\
\end{tabular}
\end{center}&&
\begin{center}
\setlength{\tabcolsep}{1pt}
  \begin{tabular}{cccccccc}
    \multicolumn{8}{c}{ Success cases from MPII and Leeds}\\
    \toprule
\includegraphics[height=0.22\linewidth]{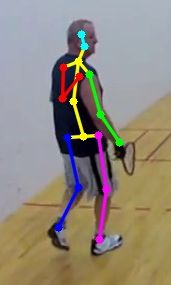}    & 
\includegraphics[height=0.22\linewidth]{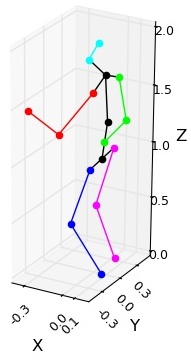} & 
\includegraphics[height=0.22\linewidth]{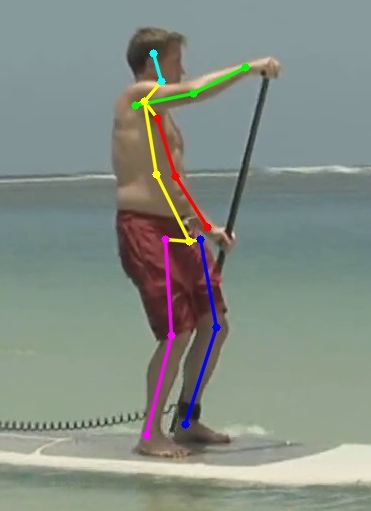}    & 
\includegraphics[height=0.22\linewidth]{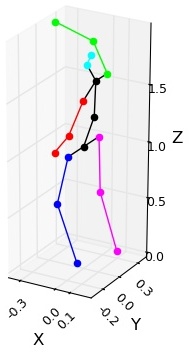} & 
\includegraphics[height=0.22\linewidth]{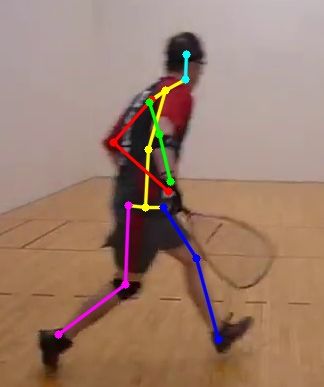}    & 
\includegraphics[height=0.22\linewidth]{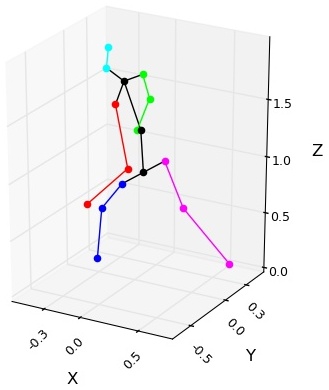} &
\includegraphics[height=0.22\linewidth]{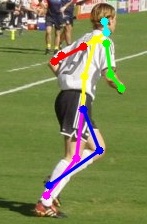}  &
\includegraphics[height=0.22\linewidth]{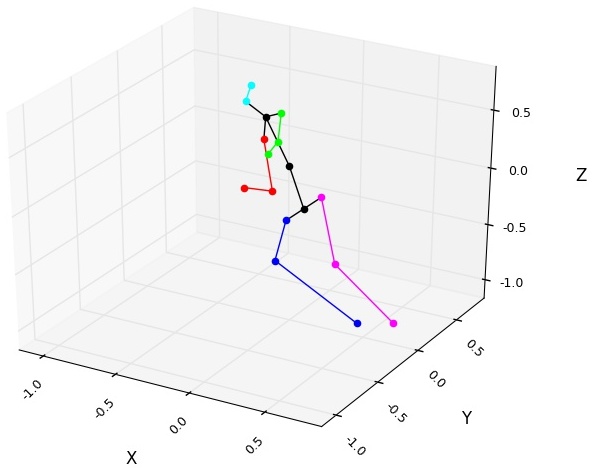} \\
\includegraphics[height=0.22\linewidth]{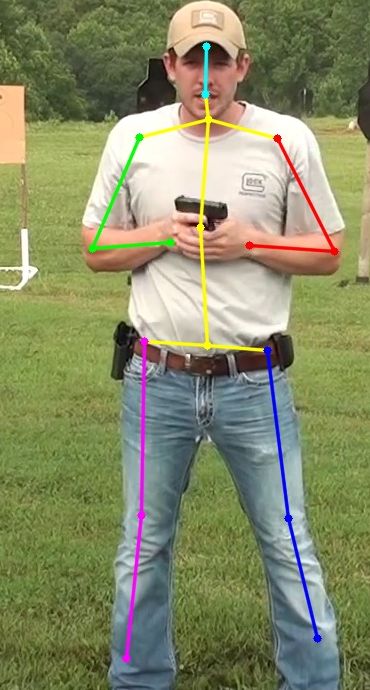}    & 
\includegraphics[height=0.22\linewidth]{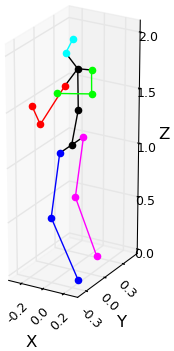} & 
\includegraphics[height=0.22\linewidth]{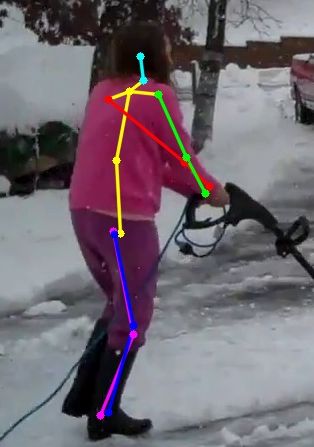}    & 
\includegraphics[height=0.22\linewidth]{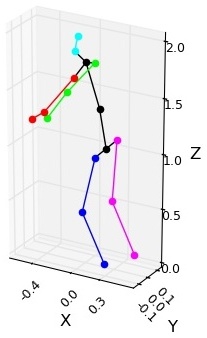} & 
\includegraphics[height=0.22\linewidth]{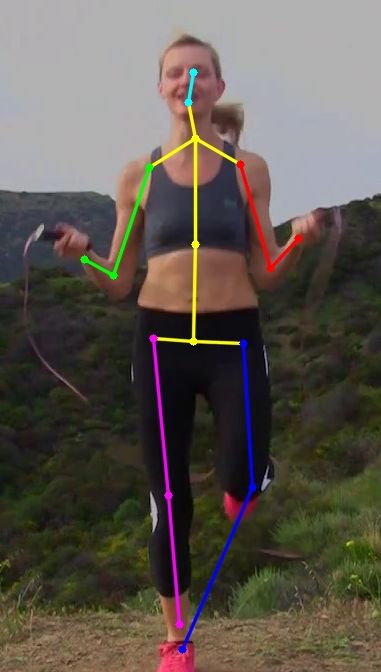}    & 
\includegraphics[height=0.22\linewidth]{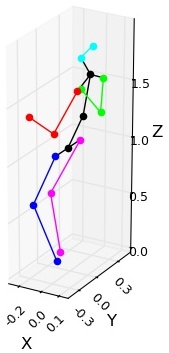} &
\includegraphics[height=0.22\linewidth]{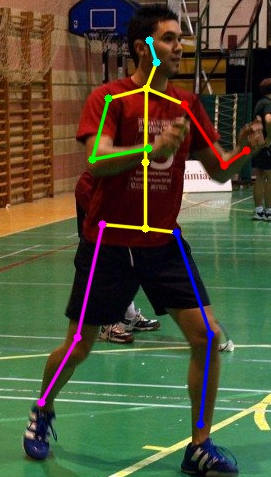} &
\includegraphics[height=0.22\linewidth]{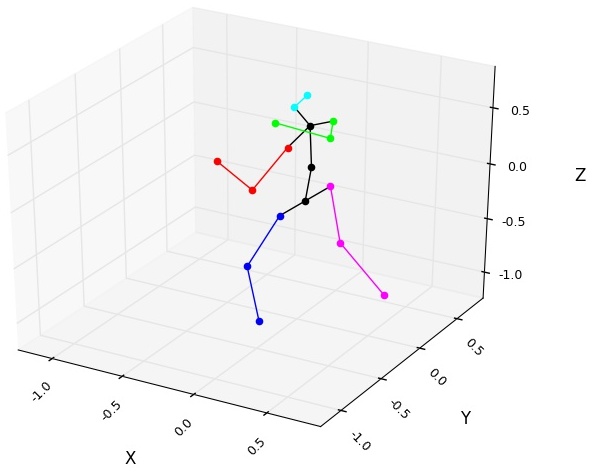} \\

\multicolumn{8}{c}{Failure cases from MPII and Leeds}\\
\toprule
\\  
\includegraphics[height=0.22\linewidth]{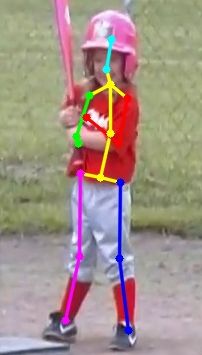}    & 
\includegraphics[height=0.22\linewidth]{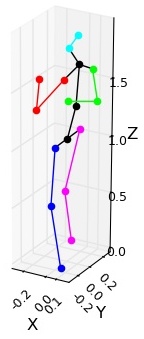} & 
\includegraphics[height=0.22\linewidth]{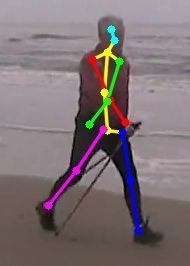}    & 
\includegraphics[height=0.22\linewidth]{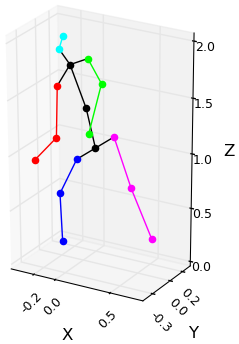} & 
\includegraphics[height=0.22\linewidth]{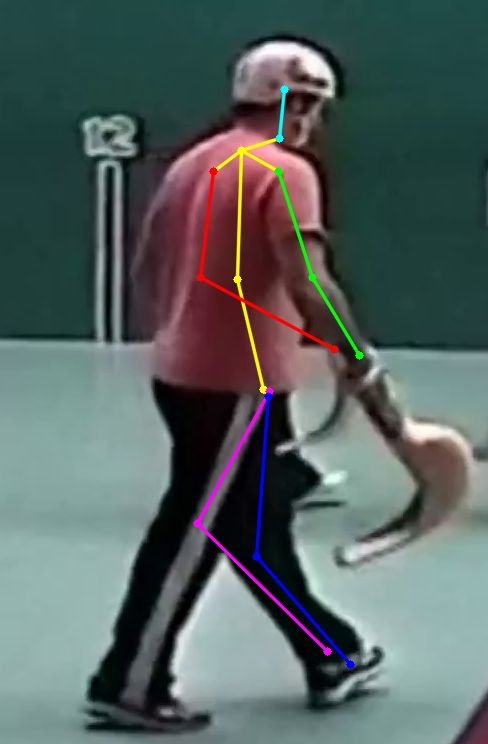}    & 
\includegraphics[height=0.22\linewidth]{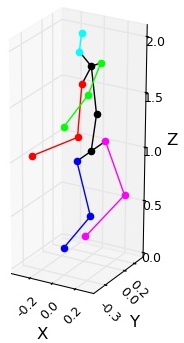} &
\includegraphics[height=0.22\linewidth]{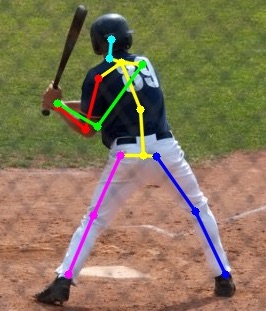}    & 
\includegraphics[height=0.22\linewidth]{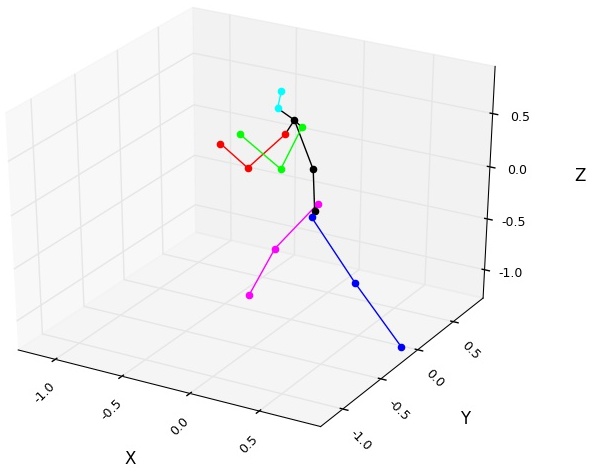} \\
\end{tabular}
\end{center}
\end{tabular}
\vspace{-5mm}
\caption{\small {\bf Left:} Results from the Human3.6M dataset. The identified 2D
  landmark positions and 3D skeleton is shown for each pose taken from
  different actions: Walking, Phoning, Greeting, Discussion, Sitting
  Down.\label{fig:Results} {\bf Right:} Results on images from the
  MPII~\cite{Andriluka:etal:CVPR:2014} (columns 1 to 3) and Leeds~\cite{Johnson10}
  datasets (last column). The  model was not trained on images as diverse as those contained in
  these datasets, however it often retrieves correct 2D and 3D joint positions. The last row
  shows example cases where the method fails either in the
  identification of 2D or 3D landmarks.\label{fig:Results_mpii}}\vspace{-5mm}
\end{figure*}
\subsection{Predicting CNN-based belief-maps}
\label{sec:CNN-belief-maps}
Convolutional Pose Machines~\cite{wei2016convolutional} can be
understood as an updating of the earlier work of
Ramakrishna~\etal~\cite{ramakrishna2014pose} to use a deep
convolutional architecture. In both approaches, at each stage $t$ and
for each landmark $p$, the algorithm returns dense per pixel belief
maps $\mathbf{b}_{t}^{p}[u,v]$, which show how confident it is that a
joint center or landmark occurs in any given pixel $(u,v)$.
For stages $t\in\{2,\ldots,T\}$ the belief maps are a function of
not just the information contained in the image but also the
information computed by the previous stage.

In the case of convolutional pose machines, and in our work which uses
the same architecture, a summary of the convolution widths and
architecture design is shown in Figure~\ref{fig:pipeline}, with
more details of training given in 
\cite{wei2016convolutional}.

Both \cite{ramakrishna2014pose,wei2016convolutional} predict the locations of
different landmarks to those captured in the Human3.6M dataset. As such the
input and output layers in each stage of the architecture are replaced with a
larger set to account for the greater number of landmarks. The new architecture
is then initialized by using the weights with those found in CPM's model for all
preexisting layers, with the new layers randomly initialized.

After retraining, CPMs return per-pixel estimates of landmark locations, while
the techniques for 3D estimation (described in the next section) make use of 2D
locations. To transform these belief maps into locations, we select the most
confident pixel as the location of each landmark
\begin{equation}
  Y_p=\argmax_{(u,v)}b_{p}[u,v]
\end{equation}
\comment{and estimate belief weights  $W_p$ as for each 2D landmark as functions of the 2D covariance
of a truncated belief map.
\begin{equation}
  W_p=\text{cov}^{-1}=\left(\sum_u\sum_v(u-Y_u^p)(v-Y_v^p)*\tilde{b}_p[u,v]\right)^{-1}
\end{equation}
where $\tilde{b}_p$ is the truncated belief map for landmark $p$ in
which all elements smaller than $3\%$ of the maximum belief are set to
zero.
}

\begin{figure}\vspace{-11mm}
    \begin{center}
      \vspace{1.2cm}
\begin{tabular}{cc}
      \multicolumn{2}{c}{2D Pose refinement in Human3.6M}\\
\includegraphics[height=0.47\linewidth]{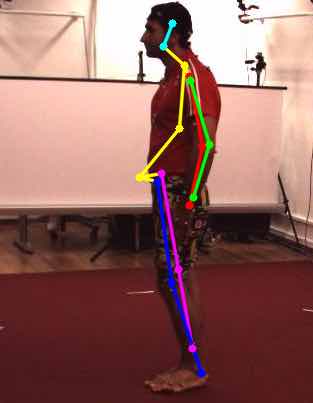}  & 
\includegraphics[height=0.47\linewidth]{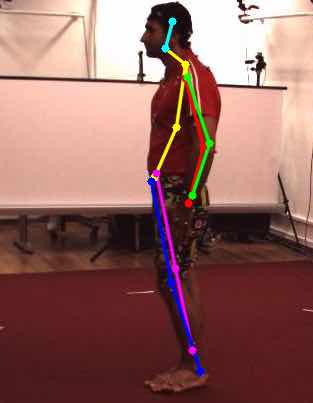}  \\
\end{tabular}
\end{center}
\vspace{-5mm}
  \caption{\small Landmark refinement:    \textit{Left}: 2D predicted landmark
    positions; \textit{Right}: improved predictions using the projected 3D
    pose.\label{fig:manifold_2D_correction}
  }
\end{figure}

\subsection{Lifting 2D belief-maps into 3D}
\label{sec:recovering-3D-from}\vspace{-2mm}
We follow~\cite{zhou2015sparseness} in assuming a weak perspective model, and
first describe the simplest case of estimating the 3D pose of a single frame
using a unimodal Gaussian 3D pose model as described in
section~\ref{sec:modeling-3D-poses}. This model is composed of a mean shape
$\mu$, a set of basis matrices ${\bf e}$ and variances $\sigma^2$, and from this
we can compute the most probable sample from the model that could give rise to a
projected image.
%
%
\begin{equation}\label{eq:get3d}
\argmin_{R,a}   ||Y -s\Pi E R (\mu + a\cdot {\bf e}  )||_2^2 +||\sigma\cdot a||_2^2
\end{equation}
Where $\Pi$ is the 
orthographic projection matrix, $E$ a
known external camera calibration matrix, and $s$ the estimated
per-frame scale. Although, given $R$ this problem is convex in $a$ and
$s$ together\footnote{To see this consider the trivial
  reparameterization where we solve for $s\mu +b \cdot {\bf e}$ and
  then let $a=b/s$.}, for an unknown rotation matrix $R$ the problem
is extremely non-convex -- even if $a$ is known -- and prone to
sticking in local minima using 
gradient
descent. Local optima often lie far apart in pose space and
 a poor optima leads to a significantly worse 3D
reconstructions.

We take advantage of the matrix $R$'s restricted form that allows it
to be parameterized in terms of a single angle $\theta$
. Rather than attempting to solve this optimization
problem using local  methods we quantize over the space
of possible rotations, and for each choice of rotation, we hold this
fixed and solve for $s$ and $a$, before picking the minimum cost
solution of any choice of $R$. With fixed choices of rotation the
terms $\Pi E R \mu$ and $\Pi E R \bf e$ can be precomputed and
finding the optimal $a$ becomes a simple linear least square problem.

This process is highly efficient and by oversampling the rotations
and exhaustively checking in $10,000$ locations we can guarantee that
a solution extremely close to the global optima is found. In
practice, using $20$ samples and refining the rotations and basis
coefficients of the best found solution using a non-linear least
squares solver obtains the same reconstruction, and we make use of the
faster option of checking $80$ locations and using the best found
solution as our 3D estimate. This puts us close to the global optima
and has the same average accuracy as finding the global
optima. Moreover, it allows us to upgrade from sparse landmark locations
to 3D using a single Gaussian at around 3,000 frames a second using
python code on a standard laptop.

To handle models consisting of a mixture of Gaussians, we follow~\cite{pitelis2013learning} and simply solve
for each Gaussian independently and select the most probably solution.

\begin{table*}[htbp]
\vspace{-5mm}
\begin{center}
  \small
\setlength\tabcolsep{3.0pt}
\begin{tabular}{lcccccccc}
\toprule
                                          & Directions     & Discussion     & Eating         & Greeting       & Phoning        & Photo           & Posing         & Purchases      \\
\toprule
LinKDE \cite{ionescu2014human3}           & 132.71         & 183.55         & 132.37         & 164.39         & 162.12         & 205.94          & 150.61         & 171.31         \\ 
Li \etal \cite{li2015maximum}             & -              & 136.88         & 96.94          & 124.74         & -              & 168.68          & -              & -              \\
Tekin \etal \cite{tekin2015predicting}    & 102.39         & 158.52         & 87.95          & 126.83         & 118.37         & 185.02          & 114.69         & 107.61         \\
Tekin \etal \cite{tekin2016structured}    & -              & 129.06         & 91.43          & 121.68         & -              & 162.17          & -              & -              \\
Tekin \etal \cite{tekin2016fusing}        & 85.03          & 108.79         & 84.38          & 98.94          & 119.39         & \textbf{95.65}  & 98.49          & 93.77          \\
Zhou \etal \cite{zhou2015sparseness}      & 87.36          & 109.31         & 87.05          & 103.16         & 116.18         & 143.32          & 106.88         & 99.78          \\
Sanzari \etal \cite{sanzari2016bayesian}  & \textbf{48.82} & \textbf{56.31} & 95.98          & \textbf{84.78} & 96.47          & 105.58          & \textbf{66.30} & 107.41         \\
\midrule
\textbf{\small Ours - Single PPCA Model}  & 68.55          & 78.27          & 77.22          & 89.05          & 91.63          & 110.05          & 74.92          & 83.71          \\
\textbf{\small Ours - Mixture PPCA Model} & 64.98          & 73.47          & \textbf{76.82} & 86.43          & \textbf{86.28} & 110.67          & 68.93          & \textbf{74.79} \\
\bottomrule
                                          & Sitting        & Sitting Down    & Smoking        & Waiting        & Walk Dog       & Walking        & Walk Together  & Average        \\
\toprule
LinKDE \cite{ionescu2014human3}           & 151.57         & 243.03          & 162.14         & 170.69         & 177.13         & 96.60          & 127.88         & 162.14         \\
Li \etal \cite{li2015maximum}             & -              & -               & -              & -              & 132.17         & 69.97          & -              & -              \\
Tekin \etal \cite{tekin2015predicting}    & 136.15         & 205.65          & 118.21         & 146.66         & 128.11         & 65.86          & 77.21          & 125.28         \\
Tekin \etal \cite{tekin2016structured}    & -              & -               & -              & -              & 130.53         & 65.75          & -              & -              \\
Tekin \etal \cite{tekin2016fusing}        & \textbf{73.76} & 170.4           & 85.08          & 116.91         & 113.72         & \textbf{62.08} & 94.83          & 100.08         \\
Zhou \etal \cite{zhou2015sparseness}      & 124.52         & 199.23          & 107.42         & 118.09         & 114.23         & 79.39          & 97.70          & 113.01         \\
Sanzari \etal \cite{sanzari2016bayesian}  & 116.89         & \textbf{129.63} & 97.84          & \textbf{65.94} & 130.46         & 92.58          & 102.21         & 93.15          \\
\midrule
\textbf{\small Ours - Single PPCA Model}  & 115.94         & 185.72          & 88.25          & 88.73          & 92.37          & 76.48          & 77.95          & 92.96          \\
\textbf{\small Ours - Mixture PPCA Model} & 110.19         & 173.91          & \textbf{84.95} & 85.78          & \textbf{86.26} & 71.36          & \textbf{73.14} & \textbf{88.39} \\
\bottomrule
\end{tabular}
\end{center}\vspace{-5mm}
\caption {\small A comparison of the 3D pose estimation results of our
  approach on the Human3.6M dataset against competitors that follow
  \textit{Protocol \#1} for evaluation (3D errors are given in mm). We
  substantially outperform all other methods in terms of average error
  showing a 4.7mm average improvement over our closest
  competitor. Note that some
  approaches~\cite{tekin2015predicting,zhou2015sparseness} use video
  as input instead of a single frame.\label{tab:comparison_1}}
\end{table*}

\subsection{Projecting 3D poses onto 2D belief maps}\vspace{-2mm}
\label{sec:projecting_from_3D_to_2D}
The  \textit{projected pose model} is interleaved throughout the
architecture (see Figure~\ref{fig:pipeline}). The goal is to correct
the beliefs regarding landmark locations at each stage, by fusing
extra information about 3D physical plausibility. Given the solution
$R$, $s$, and $a$ from the previous component, we estimate a physically
plausible projected 3D pose as
\begin{equation}\label{eq:proj}
  \hat{Y} = s\Pi E R (\mu + a\cdot {\bf e}  )
\end{equation}
which is then embedded in a belief map as
\begin{equation}
  \hat b^p_{i,j}=\begin{cases}
    1                           & \text{if} (i,j)=\hat Y_p                                                            \\
    0                           & \text{otherwise.}
  \end{cases}
\end{equation}
and then convolved using Gaussian filters.

\subsection{2D Fusion of belief maps}\vspace{-2mm}
\label{sec:fusion_layer}
The 2D belief maps predicted by the  \emph{probabilistic 3D pose model}
are fused with the CNN-based belief maps  $b^p$ according
to the following equation 
\begin{equation}
f_{t}^{p} =  w_t*b_t^p+(1 - w_t)*\hat{b}_t^p
\end{equation}
where $w_t\in[0,1]$ is a weight trained as part of the end-to-end
learning. This set of fused belief maps $f_t$ is then passed to the next
stage and used as an input to guide the 2D re-estimation of joint
locations, instead of the belief maps $b_t$ used by convolutional pose machines.
%

\subsection{The Objective and Training}
Following \cite{wei2016convolutional}, the objective or cost function $c_t$
minimized at each stage is the the squared distance between the generated fusion
maps of the layer $f^p_t$, and ground-truth belief maps $b_*^p$ generated by
Gaussian blurring the sparse ground-truth locations of each landmark $p$
\begin{equation}
c_t=\sum_{p=1}^{L+1}\sum_{z \in Z} ||f_t^p -b_*^p||_2^2
\end{equation}
For end-to-end training the total loss is the sum over all layers $\sum_{t\leq
  6} c_t$. The novel layers were implemented as an extension of the published
code of Convolutional Pose Machines~\cite{wei2016convolutional} inside the
\textit{Caffe} framework~\cite{jia2014caffe} as Python layers, with weights
updated using Stochastic Gradient Descent with momentum. Details of the novel
gradient updates used lifting estimates through 3d pose space are given in the
supplementary materials.

\section{Experimental evaluation}
\label{sec:exper-eval}
\paragraph{Human3.6M dataset:}
The  model was trained and tested on the Human3.6M dataset consisting
of 3.6 million accurate 3D human poses \cite{ionescu2014human3}. This
is a video and mocap dataset of 5 female and 6 male subjects, captured from 4
different viewpoints, that show them performing typical activities
(talking on the phone, walking, greeting, eating, etc.).

\noindent{\bf 2D Evaluation:}
Figure~\ref{fig:manifold_2D_correction} shows how the 2D predictions
are improved by the  projected pose model, reducing the overall mean
error per landmark. The 2D error reduction  using our full approach
over the estimates of~\cite{wei2016convolutional} is comparable in
magnitude to the improvement due to the change of architecture moving
from the work Zhou~\etal\cite{zhou2015sparseness} to the
state-of-the-art 2d architecture~\cite{wei2016convolutional} (i.e. a
reduction of 0.59 pixels vs. 0.81 pixels). See
Table~\ref{tab:2d_comparison} for details.

\begin{table}[h]
  \begin{center}
    \small
\begin{tabular}{lc}
    \toprule
  Evaluation of 3D error (mm)& Protocol \#2\\
  \midrule
  Yasin~\etal~\cite{Yasin:etal:CVPR:2016} & 108.3\\
  Rogez~\etal~\cite{rogez2016mocap} & 88.1\\
  \textbf{Ours - Mixture PPCA Model} & \textbf{70.7}\\
  \\
  \toprule
  Evaluation of 3D error (mm)& Protocol \#3\\
  \midrule
  Bogo~\etal~\cite{bogo2016keep} & 82.3\\
  \textbf{Ours - Mixture PPCA Model} & \textbf{79.6}\\
  \\
    \toprule
  Evaluation of 2D pixel error& \\
  \toprule
  Zhou \etal~\cite{zhou2015sparseness} & 10.85\\
  \midrule
  Trained CPM~\cite{wei2016convolutional} architecture   & 10.04\\
  {\bf Ours} using 3D refinement & \textbf{9.47}\\
\end{tabular}
\end{center}\vspace{-5mm}
\caption {\small Further evaluation on the Human3.6M dataset. Top two
  tables compare our 3D pose estimation errors against competitors
  on \textit{Protocols \#2} or
  \textit{\#3}. \label{tab:comparison_2}\label{tab:comparison_3}
  Bottom table compares our 2D pose estimation error against
  competitors. Our approach, which lifts the 2D landmark predictions
  into a plausible 3D model and then projects them back into the
  image, substantially reduces the error. Note
  that~\cite{zhou2015sparseness} use video as input and knowledge of
  the action label.\label{tab:2d_comparison}\vspace{-8mm}}
\end{table}

\noindent{\bf 3D Evaluation:} Several evaluation protocols have been followed by
different authors to measure the performance of their 3D pose estimation methods
on the Human3.6M dataset. Tables~\ref{tab:comparison_1} and
~\ref{tab:comparison_2} show comparisons of the 3D pose estimation with previous
works, where we take care to evaluate using the appropriate protocol.

\textit{{\bf Protocol \#1}}, the most standard evaluation protocol on Human3.6M,
was followed by~\cite{ionescu2014human3, li2015maximum, tekin2015predicting,
  tekin2016structured, tekin2016fusing, zhou2015sparseness,
  sanzari2016bayesian}. The training set consists of 5 subjects (S1, S5, S6, S7,
S8), while the test set includes 2 subjects (S9, S11). The original frame rate
of 50 FPS is down-sampled to 10 FPS and the evaluation is on sequences coming
from all 4 cameras and all trials. The reported error metric is the \emph{3D
  error} i.e. the Euclidean distance from the estimated 3D joints to the ground
truth, averaged over all 17 joints of the Human3.6M skeletal model.
Table~\ref{tab:comparison_1} shows a comparison between our approach and
competing approaches using \textit{Protocol \#1}.
Our baseline method using a single unimodal probabilistic PCA model outperforms
almost every method in most action types, with the exception of
Sanzari~\etal~\cite{sanzari2016bayesian}, which it still outperforms on average
across the entire dataset. The mixture model improves on this again, offering a
$4.76$mm improvement over Sanzari~\etal, our closest competitor.

\textit{{\bf Protocol \#2}}, followed
by~\cite{Yasin:etal:CVPR:2016,rogez2016mocap}, selects 6 subjects (S1, S5, S6,
S7, S8 and S9) for training and subject S11 for testing. The original video is
down-sampled to every 64th frame and evaluation is performed on sequences from
all 4 cameras and all trials. The error metric reported in this case is
the \emph{3D pose error} equivalent to the per-joint 3D error up to a similarity
transformation (i.e. each estimated 3D pose is aligned with the ground truth
pose, on a per-frame basis, using Procrustes analysis). The error is averaged
over 14 joints.  Table~\ref{tab:comparison_2} shows a comparison between our
approach and other approaches that use \textit{Protocol \#2}. Although, our
model was trained using only the 5 subjects used for training in \emph{Protocol
  \#1} (one fewer subject), it still outperforms the other
methods~\cite{rogez2016mocap,Yasin:etal:CVPR:2016}.

\textit{{\bf Protocol \#3}}, followed by \cite{bogo2016keep}, selects the same
subjects for training and testing as \emph{Protocol \#1}. However, evaluation is
only on sequences captured from the frontal camera (``cam 3'') from trial 1 and
the original video is not subsampled. The error metric used in this case is the \emph{3D
  pose error} as described in Protocol \#2. The error is averaged over a subset
of 14 joints. Table~\ref{tab:comparison_3} shows a comparison between our
approach and \cite{bogo2016keep}. Our method outperforms
Bogo~\etal~\cite{bogo2016keep} by almost 3mm on average, even though Bogo~\etal
exploits a high-quality detailed statistical 3D body model~\cite{loper2015smpl}
trained on thousands of 3D body scans, that captures both the variation of human
body shape and its deformation through pose.

\comment{\textcolor{red}{REMOVE IF NOT COMPARING WITH M.BLACK. For the
    sake of comparison, a subset of the test-set has been used to
    compare the performance of our model with methods that used a
    different convention~\cite{akhter2015pose}. In this case the test
    set consisting of subjects (S9, S11) using frames captured by
    \textit{camera 3} only, and considering for each action just
    \textit{trial 1}. Frames have been sampled at 10 FPS.}}

\paragraph{MPII and Leeds datasets:}
The  proposed approach trained exclusively on the Human3.6M dataset can
be used to identify 2D and 3D landmarks of images contained in
different datasets. Figure~\ref{fig:Results_mpii} shows some
qualitative results on the MPII
dataset~\cite{Andriluka:etal:CVPR:2014} and
on the Leeds dataset ~\cite{Johnson10}, including failure cases. Notice how the 
\textit{probabilistic 3D pose model} generates anatomically plausible
poses even though the 2D landmark estimations are not all
correct. However, as shown in bottom row, even small errors in 2D pose
can lead to drastically different 3D poses. These inaccuracies could
be mitigated without further 3D data by annotating additional RGB
images for training from different datasets.

\section{Conclusion}
We have presented a novel approach to human 3D pose estimation from a
single image that outperforms previous solutions. We approach this as
a problem of iterative refinement in which 3D proposals help refine
and improve upon the 2D estimates.
Our approach shows the importance of thinking in 3D even for 2D pose
estimation within a single image, with our method 
demonstrating better 2D accuracy than~\cite{wei2016convolutional}, the 2D approach it is
based upon. 
Our novel approach for upgrading from 2D to 3D is extremely efficient. When
using 3 models, as in Tables \ref{tab:comparison_1} and \ref{tab:comparison_2},
the upgrade for each stage in CPU-based Python code runs at approximately 1,000
frames a second, while a GPU-based real-time approach for Convolutional Pose
Machines has been announced. Integrating these systems to provide a reliable
real-time 3D pose estimator is a natural future direction, as is integrating
this work with a simpler 2D approach for real-time pose estimation on lower
power devices.

\subsection*{Acknowledgments}
This work was funded by the SecondHands project, from
the European Union's Horizon 2020 Research and Innovation programme
under grant agreement No 643950.   Chris Russell was partially
supported by The Alan Turing Institute under  EPSRC grant
EP/N510129/1.

{\small
  \bibliographystyle{ieee} \bibliography{egbib} }



\section*{Supplementary material}

# Supplementary Material

## 1. Computing derivatives for back-propagation through our lifted model

As discussed by the Convolution Pose Machine paper [1] recurrent like architectures such as ours have problems with *vanishing gradients* and for effective training they require an additional loss function to be defined for each layer, that independently drives each individual layer to return correct predictions regardless of how this information is used in subsequent layers.

Before we give the derivation of the gradients it should be emphasized that it is entirely possible to train the network without using them – in fact similar results can be obtained by only using the 3D lifting for the forward pass, and not back-propagating the lifting derivatives through the rest of the network. As the additional layers make use of custom Python-based derivatives rather than an efficient implementation, for computational reasons it might preferable to avoid this step. Nonetheless for completeness we include the derivatives.

There are two reasons the gradients are unneeded: Our lifting 3D model we use makes its best predictions when the 2D predictions of the same layer are closest to ground truth, and this is a constraint naturally enforced by the objective of equation (8) of the main paper. Further, as with Convolutional Pose Machines [1] our architecture suffers from problems with vanishing gradients. To overcome this Wei *et al*. [1] defined an objective at each layer, which acted to locally strengthen the gradients. However, a side effect of this multi-stage objective is that most of the effects of back-propagation happen locally and gradients back-propagated from other layers have little effect on the learning. This makes subtle interactions between layers less influential, and forces the learning process to concentrate on simply making accurate 2D predictions in each layer.

We first give the results for computing the gradients of sparse predicted locations $\hat{Y}$ from $Y$ (see section 5 of the main paper), before discussing the gradients induced on the confidence maps by these sparse locations.

### 1.1. Landmark Gradients

In the interests of readability we neglect the use of indices to indicate stages, the reader should assume that all variables are taken from the same stage. Similarly, when dealing with a mixture of Gaussians, as we are only interested in computing a sub-gradient, the reader should assume that the best model has already been selected in the forward pass and we are computing gradients using only this model.

Recall (section 5 of main body of paper) that the mapping from the initial landmarks $Y$ to the projected 3D proposals $\hat{Y}$ is given by

$$\hat{Y} = \Pi \mathbf{R}(\mu + a \cdot \mathbf{e}) \tag{1}$$

where

$$\mathbf{R}, a = \underset{\{\mathbf{R}^* \in \mathcal{R}, a^* \in \mathbb{R}^J\}}{\arg\min} ||Y - \Pi \mathbf{R}^*(\mu + a^* \cdot \mathbf{e})||_2^2 + (\sigma \cdot a^*)^2 \tag{2}$$

where $\mathcal{R}$ is a discrete set of rotations we exhaustively minimize over, and $J$ is the number of bases in $\mathbf{e}$. Owing to the use of discrete rotations, this mapping from $Y$ to $\hat{Y}$ is a piecewise smooth approximation of the smooth function defined over a continious $\mathcal{R}$, and sub-gradients can be induced by fixing $R$ to its current state. Hence:

$$\frac{d\hat{Y}}{da} = \Pi \mathbf{R} \mathbf{e} \tag{3}$$

For the remainder of the section, and to compact notation we will write $E$ for the matrix of size $2L \times J$ ($L$ the number of landmark points and $J$ being the number of bases in $\mathbf{e}$ ) formed by unwrapping tensor $\Pi \mathbf{R} \mathbf{e}$. Similarly, we will unwrap the $2 \times L$ matrices $Y$ and $\hat{Y}$ and write them as $y$ and $\hat{y}$. We also write $p$ for the vector representing the unwrapped set of 2D landmark positions $\Pi \mathbf{R} \mu$.

We will use $[y, 0]$ for the vector formed by vector $y$ followed by $J$ zeros, and $\bar{E}$ for the matrix of size $(2L + J) \times J$ formed by concatenating $E$ with the matrix that has values $\sigma$ along the diagonals and zero everywhere else. We can rewrite equation (3) in its new notation as:

$$\frac{d\hat{y}}{da} = E \tag{4}$$

and given $R$, we can rewrite equation (2) as

$$a = \underset{a* \in \mathbb{R}^J}{\arg\min} \, |[y, 0] - [p, 0] - a^* \bar{E}\|_2^2 \tag{5}$$

or

$$a = [(y - p), 0]\bar{E}^\dagger \tag{6}$$

with $\bar{E}^\dagger$ continuing to represent the pseudo-inverse of $\bar{E}$. Hence

$$\frac{da}{d[y, 0]} = \bar{E}^\dagger \tag{7}$$

and

$$\frac{d\hat{y}}{dy} = \frac{d\hat{y}}{da}\frac{da}{dy} = E\bar{E}^t \tag{8}$$

where $\bar{E}^t$ is the truncation of $\bar{E}^\dagger$.

## 1.2. Mapping belief gradients to coordinate transform

The coordinates of each predicted landmark $\hat{Y}_p$ induce a Gaussian in the belief map $\hat{b}_p$. So a change in the $x$ component of $\hat{Y}_p$ induces an update which is equivalent to a difference of Gaussians.

$$\frac{d\hat{b}_p}{d\hat{Y}_p^x} \approx \frac{G(\hat{Y}_p^{(x)} + \delta_x) - G(\hat{Y}_p^{(x)} - \delta_x)}{2\delta_x} \tag{9}$$

and the same for the $y$ component as well. For computational purposes we take $\delta_x$ as one pixel. As such, an induced gradient on the projected belief map near the predicted location $\hat{Y}$ $\hat{b}_p$ induces an updating of $\hat{Y}$ that is propagated through to $Y$ using the sub-gradients described in equation (8).

**Updating $B$**  Writing $B$ for the the set of all $b_p$, and assuming $Y_p$ is not in the right location, i.e. given updates $\Delta \hat{B}$ on $\hat{B}$ such that

$$\Delta \hat{B} \cdot \frac{d\hat{B}}{d\hat{Y}} \frac{d\hat{Y}}{dY_p} \neq 0,$$

any update of $b$ in which we decrease the belief at $b_{Y_p}^p$ and increase anywhere else is a valid sub-gradient. We choose as a sensible update a negative step at $b_p$ of magnitude $m = k\|\Delta \hat{B} \cdot \frac{d\hat{B}}{d\hat{Y}} \frac{d\hat{Y}}{dY_p}\|$ and a positive update for each element $Y$ of of $B_p$ of the magnitude $m \cdot N(Y, \sigma^2)$ in the quadrant of a Gaussian of the same width used to generate $\hat{b}$ (i.e. $\sigma = 1$ see section 5.6 of main paper) and with the same direction as $\Delta \hat{B} \cdot \frac{d\hat{B}}{d\hat{Y}} \frac{d\hat{Y}}{dY_p}$ in each $x$ and $y$ coordinate.



\end{document}